\pdfoutput=1

\documentclass[11pt]{article}

\usepackage{acl}

\usepackage{times}
\usepackage{latexsym}
\usepackage{hhline}
\usepackage{xspace}
\usepackage{amsmath,amsfonts,amssymb}
\usepackage{url}
\usepackage[noend,ruled,linesnumbered,vlined]{algorithm2e}

\usepackage[T1]{fontenc}

\usepackage[utf8]{inputenc}

\usepackage{microtype}

\definecolor{light_blue}{HTML}{DCE6F1}
\usepackage[most]{tcolorbox}
\tcbset{on line, 
        boxsep=1pt, left=0pt,right=0pt,top=0pt,bottom=0pt,
        colframe=white,colback=light_blue,  
        highlight math style={enhanced}
        }

\usepackage{caption}
\usepackage{graphicx}
\usepackage{booktabs}
\usepackage{xcolor}
\usepackage{multirow}
\usepackage{subcaption}
\usepackage[shortlabels]{enumitem}
\usepackage{xspace}
\usepackage{soul}

\DeclareMathOperator*{\argmax}{arg\,max}

\definecolor{myred}{rgb}{0.85, 0.25, 0.25}
\definecolor{myblue}{rgb}{0.3, 0.3, 0.85}

\newcommand\model{\textsc{QAMDen}\xspace}
\newcommand\primera{\textsc{Primera}}
\newcommand\pegasus{\textsc{Pegasus}}
\newcommand\cdlm{\textsc{CDLM}}
\newcommand\qasem{\textsc{QASem}}

\title{\textit{Peek Across:} Improving Multi-Document Modeling \\ via Cross-Document Question-Answering}

\author{Avi Caciularu$^{1}\thanks{\;\; Work partly done as an intern at AI2.}$\hspace{1em} Matthew E. Peters$^{2}$\hspace{1em} Jacob Goldberger$^1$\\
\textbf{Ido Dagan$^1$\hspace{1em} Arman Cohan$^{2,3}$} \vspace{6pt}\\  
    $^1$Bar-Ilan University, Ramat-Gan, Israel\\
    $^2$Allen Institute for AI, Seattle, WA\\
    $^3$Yale University, New Haven, CT\\
    {\small\tt avi.c33@gmail.com, arman.cohan@yale.edu}
}
\begin{document}
\maketitle

\begin{abstract}
The integration of multi-document pre-training objectives into language models has resulted in remarkable improvements in multi-document downstream tasks. 
In this work, we propose extending this idea by pre-training a generic multi-document model from a novel cross-document question answering pre-training objective.
To that end, given a set (or cluster) of topically-related documents, we systematically generate semantically-oriented questions from a salient sentence in one document and challenge the model, during pre-training, to answer these questions while ``peeking" into other topically-related documents.
In a similar manner, the model is also challenged to recover the sentence from which the question was generated, again while leveraging cross-document information.
This novel multi-document QA formulation directs the model to better recover cross-text informational relations, and introduces a natural augmentation that artificially increases the pre-training data. 
Further, unlike prior multi-document models that focus on either classification or summarization tasks, our pre-training objective formulation enables the model to perform tasks that involve \emph{both} short text generation (e.g., QA) and long text generation (e.g., summarization).
Following this scheme, we pre-train our model -- termed \model{} -- and evaluate its performance across several multi-document tasks, including multi-document QA, summarization, and query-focused summarization, yielding improvements of up to 7\%, and significantly outperforms zero-shot GPT-3.5 and GPT-4.\footnote{Our code is available at \url{https://github.com/aviclu/peekacross}.}
\end{abstract}

\section{Introduction}
\label{sec:intro}

Among recent NLP research, multi-document processing is gaining increasing attention, due to the need to handle and process an increasing amount of textual data and available documents online.
A number of prominent applications that are concerned with aggregating information from multiple texts are multi-document summarization~\cite{fabbri-etal-2019-multi,10.1145/3397271.3401327}, query-focused multi-document summarization~\cite{xu-lapata-2020-coarse,pasunuru2021data}, and multi-hop question answering~\cite{yang-etal-2018-hotpotqa,welbl-etal-2018-constructing}. These tasks remain challenging mostly since existing NLP models are designed to handle single texts, rather than processing multiple documents at once~\cite{caciularu-etal-2021-cdlm-cross}. 

\begin{figure}[t]
    \centering
    \includegraphics[scale=0.6]{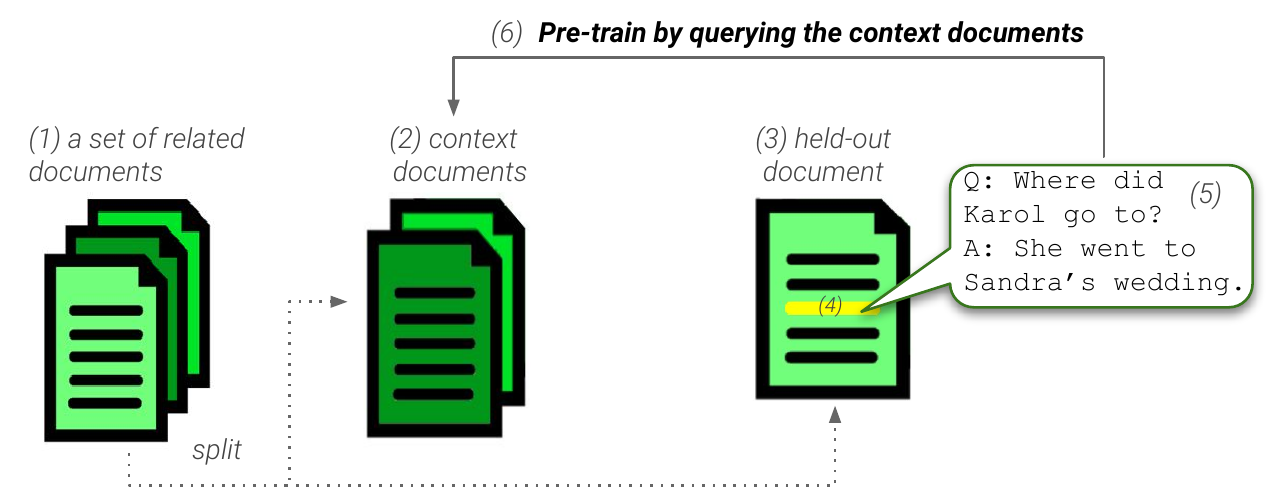}
    \caption{Illustration of our pre-training and data generation. Per a considered set of related documents (1) which we split into \emph{context documents} (2) and a \emph{held-out document} (3), we select the most salient sentence (4) that is used for generating a question-answer pair (5). Then, we pre-train a model by generating the proper answer and the salient sentence, given the question and the context documents (6).}
    \label{figure:intro}
\end{figure}

Early solutions for multi-text processing were task-specific and used complex architectures that were difficult to generalize across different multi-document tasks~\cite{liu-lapata-2019-hierarchical,wang-etal-2020-heterogeneous,ginzburg-etal-2021-self}. Efficient LMs~\cite{tay2021long,beltagy2020longformer} recently demonstrated that by simply concatenating multiple documents into a single sequence, the transformer can offload the goal of identifying and connecting relevant information between the documents. Recently, it was suggested that these long-context LMs can be equipped with new pre-training objectives to enable them to process multiple documents more effectively~\cite{caciularu-etal-2021-cdlm-cross, xiao-etal-2022-primera, yasunaga-etal-2022-linkbert}.

These pre-trained models demonstrated state-of-the-art performance on a variety of multi-document downstream tasks, and outperformed underlying LMs and task-specific architectures. 
Such models are often pre-trained using a dataset where each instance is a set of related documents (e.g., news articles all discussing a specific event), which facilitates modeling of cross-text relationships.
Existing multi-document pre-training objectives involve unmasking tokens in a document \cite{caciularu-etal-2021-cdlm-cross}, or generating a salient masked sentence \cite{pegasus,xiao-etal-2022-primera}, encouraging the model to recover missing information using other documents. While successful, these models are either limited to classification tasks \cite{caciularu-etal-2021-cdlm-cross} or primarily designed for summarization \cite{pegasus,xiao-etal-2022-primera}. 

In this work, we propose a novel pre-training objective that supports both short and long text generation, resulting in a versatile and general multi-document language model. In particular, we hypothesize that using questions and answers involving multiple documents can encourage the model to better learn and incorporate both fine-grained information (by asking questions about core
information units in a specific sentence) as well as coarse-grained cross-document relationships required to generate a long text such as a summary. We show that this approach holds not only for summarization, but for other multi-document downstream tasks as well.

During the pre-training of existing multi-document language models, the goal is to unmask spans (for encoder-only models) or generate masked textual spans (for encoder-decoder models) under a multi-document context. To that end, multiple concatenated sequences of related documents are fed during pre-training, thus requiring a large number of sets of related documents for an effective pre-training phase~\cite{hoffmann2022training}. In a variety of existing multi-document benchmarks, such as multi-document summarization, only small to medium-scale document clusters are readily available. These are acquired either automatically with lexical similarity and retrieval~\cite{fabbri-etal-2019-multi} or semi-automatically~\cite{newshead}, but generally, this process requires a substantial amount of human effort for filtering instances and generating high quality corpora.

By employing a novel multi-document question-answer generation procedure, we propose an effective method for expanding the multi-document pre-training corpora. Our approach allows us to provide multiple views for every single cluster of documents, thereby artificially increasing the pre-training data size (in terms of number of instances) via augmentation. 
To expose the model to a variety of contexts and diversify the pre-training data, we propose to generate multiple pairs of questions and answers and condition them on a subset of the documents' cluster. We select a salient sentence in one \emph{held-out}
document and then employ a recent parser to generate a high-quality question-answer pair about one predicate in the selected sentence, using a systematic semantically-oriented approach~\cite{Klein2022QASemPT}. This new multi-document pre-training objective challenges the model to generate both the answer to the question as well as the salient sentence, while discarding the held-out document or parts of it (see Figures~\ref{figure:intro}, \ref{figure:data} for illustration). This procedure exposes the model to a variety of contexts  -- a question and a different subset of the documents in the cluster per instance, in contrast to prior methods that provide only a single view of the cluster. 
Our contributions are summarized below:
\begin{itemize}[noitemsep,topsep=0pt]
    \item A new pre-training approach for multi-document modeling, formulated as a cross-document question answering task, further directing the LM to model cross-text relationships, focusing on both fine- and coarse-grained information. 
    \item The number of pre-training examples generated by our suggested method is not bounded by the number of clusters, allowing the production of a variety of cross-document contexts.
    \item The resulting \underline{Q}uestion-\underline{A}nswering-based \underline{M}ulti-\underline{D}ocum\underline{EN}t (\model{}) model advances the state-of-the-art for several multi-document tasks.
\end{itemize}
\section{Related Work}
\label{sec:related_work}

Long-context efficient text generation transformers~\cite{tay2021long,long-range-transformers} extend earlier transformer models \cite{vaswani2017attention} for processing long sequences, often using a sparse self-attention architecture. Examples include the Longformer Encoder-Decoder (LED)~\cite{beltagy2020longformer}, and LongT5~\cite{guo-etal-2022-longt5}. These models demonstrated that single-text approaches be can adapted to multi-document tasks by concatenating multiple documents into a single sequence and processing them using their sparse attention patterns. They sparsify the full self-attention matrix of transformers by using a combination of a localized sliding window (called local attention), as well as a global attention pattern on a few specific input locations. LED is build upon the BART model~\cite{lewis-etal-2020-bart} by using additional positional embeddings and global attention weights, and introduces the global attention mode that operates over pre-selected tokens. LongT5 extends the T5 model~\cite{T5} by using a similar technique introduced in the ETC and \textsc{BigBird} models~\cite{ainslie-etal-2020-etc,zaheer2020bigbird}, relieving the requirement to manually select global tokens by automatically globalizing the aggregated representations of groups of tokens. 

Further strategies have been proposed for increasing these models' abilities in multi-document tasks. The Cross-Document Language Model (CDLM)~\cite{caciularu-etal-2021-cdlm-cross} suggested pre-training a Longformer-encoder~\cite{beltagy2020longformer} over sets of related documents, and showed superior performance results over several multi-document tasks. Following this methodology, the authors of LinkBERT~\cite{yasunaga-etal-2022-linkbert} used a similar approach, but utilized Wikipedia's hyperlinks in order to curate informative pairs of linked documents for LM pre-training.

In order to adopt the multi-document pre-training approach for sequence-to-sequence tasks, \primera{}~\cite{xiao-etal-2022-primera}, which is built on top of the Longformer encoder-decoder model (LED), selected salient sentences within clusters of related documents using a pyramid estimation approach, resembling the method presented for pre-training the single-document \pegasus{} model~\cite{pegasus}. While this work is the closest to ours, it was pre-trained to generate masked salient sentences without any control, which makes the model potentially hallucinate while generating text, while our model uses a controlled QA-based objective. Furthermore, unlike these works, our method generates significantly more data then used to pre-train \primera{}, which is possible to obtain by the single-document QA generation approach. Our QA pre-training formulation allows us to generate multiple contexts per document cluster. 

Another related line of work includes methods that incorporate large-scale QA-generated data for pre-training LMs~\cite{he-etal-2020-quase,jia-etal-2022-question,huber-etal-2022-ccqa}. These works hypothesize and show that pre-training by utilizing generated QA data can encourage contextual representations to encode useful semantic information for other non-QA downstream tasks. Inspired by that, we conjecture that LMs can strongly benefit from infusing QA during pre-training in the multi-document setup, for adding an additional signal for modelling cross-text relationships.
\section{Augmenting the Multi-Document Pre-training objective}
\label{sec:data_creation}

\begin{figure}[t]
    \centering
    \includegraphics[scale=.55]{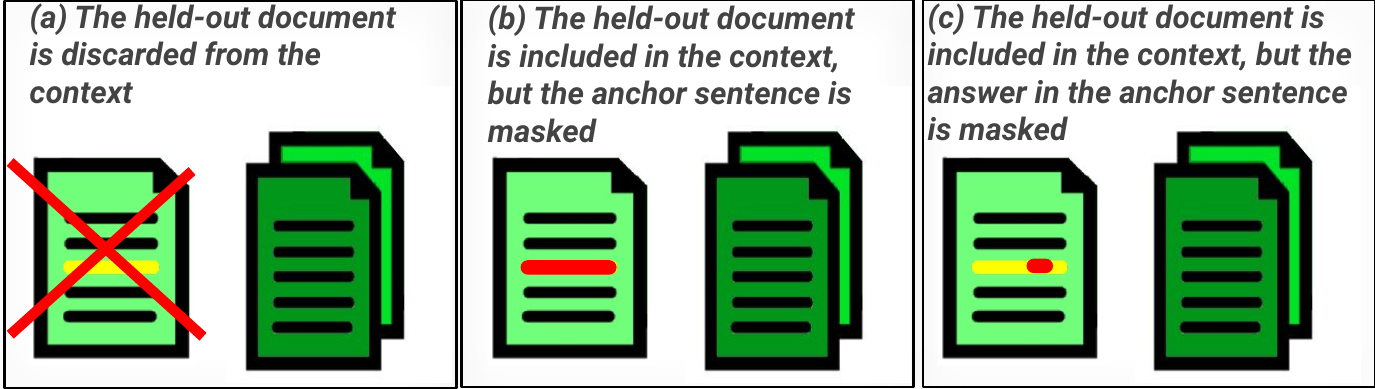}
    \caption{A schematic of our pretraining data modes. The salient sentence which is used for QA generation is colored in yellow. (a) The context does not include the held-out document, therefore this mode is the most challenging. (b) The held-out document is present in the context, but the salient sentence used for the QA generation is masked (red). (c) The held-out document is present in the context, but the answer span within the salient sentence is masked (red).}
    \label{figure:data}
\end{figure}

In this section, we provide the required steps for compiling the pre-training dataset for \model{}. We next elaborate on the details of the data creation and provide analysis of the resulted corpus. 

Recent works have shown that for text summarization, pre-training LMs to generate a ``summary-like'' sequence, termed \textit{pseudo summary}, inherently provides gains over general-purpose pre-trained LMs \citep[\pegasus, \primera;][]{pegasus,xiao-etal-2022-primera}. The data in which the \pegasus{} and \primera{} models were pre-trained on was constructed using the Gap Sentence Generation (GSG) method, which suggests masking highly-ranked salient sentences, where salience is pre-determined by a sentence-scoring method of interest. Particularly, in \pegasus{}, GSG has been adopted as its pre-training objective, where some sentences in a \emph{single} document are masked in the input and the model is tasked to generate them.

Formally, for each sentence $s^i$ in a given input document $D$, \pegasus{} computes its salience score based on its \textsc{Rouge} score~\cite{lin-2004-rouge} w.r.t the rest of the sentences within the document ($D/\{s^i\}$), i.e. ${\mathrm{Score(}s^i) = \mathrm{\textsc{Rouge}}(s^i,D/\{s^i\})}$. 
Intuitively, this metric assigns a high score to the sentences that have a high overlap and share more lexical information with the rest of the sentences in the document, thus assigning high scores to prominent sentences.
\primera{} has generalized this notion to support the multi-document setup, by applying a GSG variant over a cluster of related documents. 

\paragraph{Cross-Document GSG.} We propose augmenting the GSG technique to formulate a cross-document question answering pre-training objective for multi-document tasks, instead of the existing pseudo summary generation methods. Our approach supports identification of both fine- and coarse-grained information as we describe below, and results in a substantially larger amount of pre-training examples compared to the preceding methods.

Formally, we are given a cluster of related documents ${\mathcal{S}=\left(D_1, D_2, \hdots, D_{|\mathcal{S}|}\right)}$ in a corpus $\mathcal{C}$. Our cross-document (CD) GSG salience score for the $i^\text{th}$ sentence within the $k^\text{th}$ document in the set ($s_k^i$), is defined by its \textsc{Rouge} score w.r.t the rest of the sentences within the document ($D_k/\{s_k^i\}$) as well as the other documents ($\mathcal{S}/D_k$), i.e. ${\text{CD-GSG-Score(}s_k^i) = \mathrm{\textsc{Rouge}}(s_k^i,\mathcal{S}/\{s_k^i\})}$. 
Then, for every document $k$, following~\citet{pegasus,xiao-etal-2022-primera} we select the top-scored sentence $s_k^*$, and then we use this sentence to generate a pair of a question and an answer. 

\begin{figure}[t]
    \centering
    \includegraphics[scale=0.38]{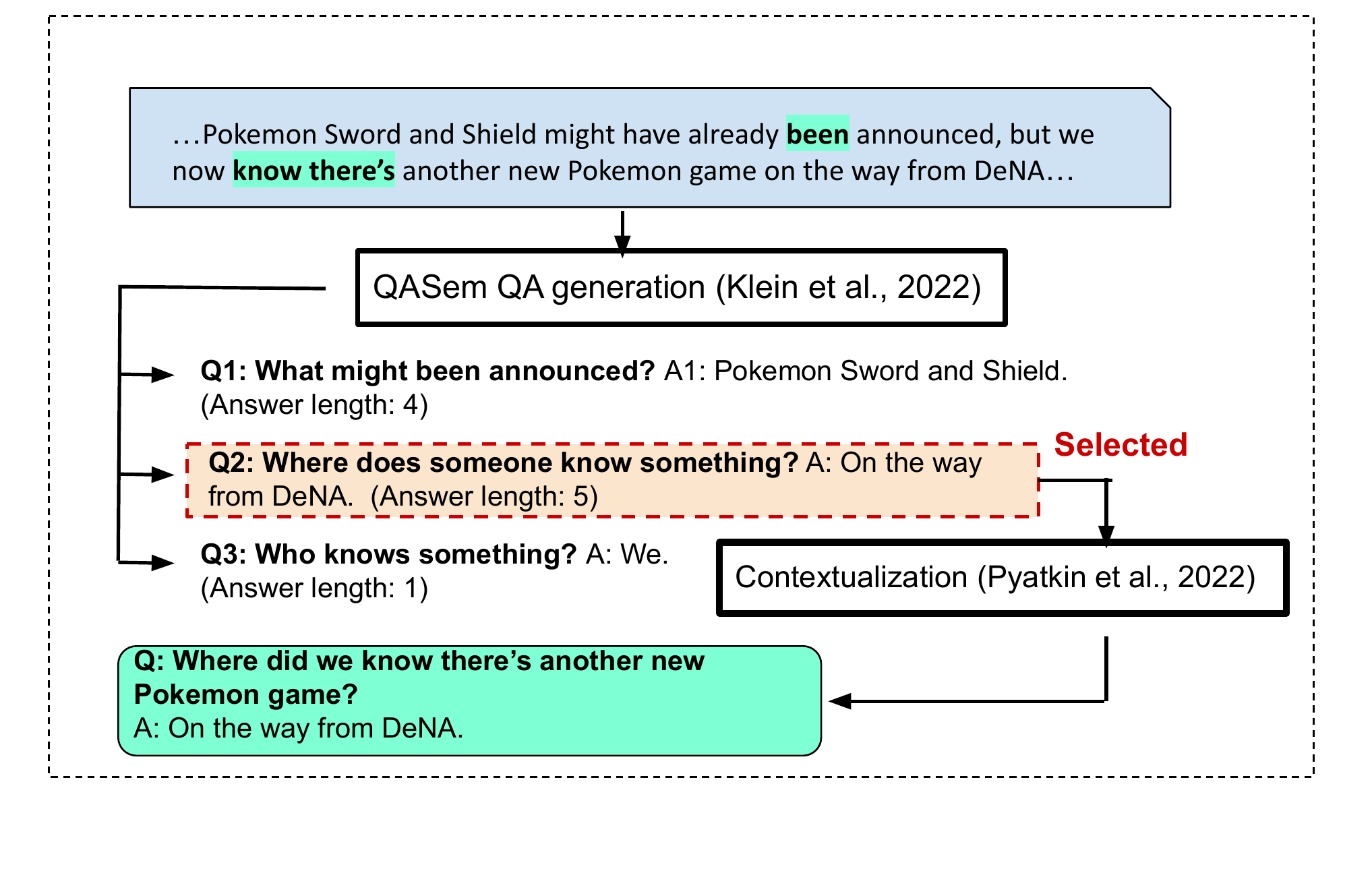}
    \caption{A schematic of the process of QA generation using \qasem{}~\cite{Klein2022QASemPT} and the contextualization model from~\citet{pyatkin-etal-2021-asking}. This is an actual sample that was created and used for pre-training \model{}, where the document is taken from NewSHead~\cite{newshead}.}
    \label{figure:qasem}
\end{figure}

\paragraph{Generating Cross-Document QAs.} 
For generating the cross-document questions and their answers, we employ \qasem{}, a recent semantic parsing framework for question generation~\cite{Klein2022QASemPT}.\footnote{We tried several leading question generation methods, and \qasem{} introduced superior quality of questions, attributed to its semi-structured nature. See \S\ref{subsec:ablation} for empirical results.} \qasem{} intended soliciting a manageable, discrete account of information in a text for the sake of building natural language semantic representations. It automatically labels each verbal predicate-argument relation with a question-answer pair, where a natural language question represents a semantic role, while the answers correspond to the arguments that appear in the input text. \qasem{} is thus an appealing approach since it is capable of generating multiple high-quality questions given a sentence. We apply \qasem{} over the sentences withing the pre-training data in order to generate question-answer pairs, and then apply the model from \citet{pyatkin-etal-2021-asking} which transforms the question into a more natural and clear form, with contextualized arguments (see example in Figure~\ref{figure:qasem}). In order to resemble a summarization task where the generated text is typically long, we select the question-answer pair with the longest argument produced by \qasem{}. Formally, \qasem{}$(\cdot)$ receives a sentence $s^*_k$ as an input, and produces question-answer pair $(q_k^*,a_k^*)$, where $a_k^*$ is the longest among the generated answers. See a detailed example and full description in App.~\ref{subsec:qasem}.

Considering the question-answer pair, our goal is to encourage the LM to generate the correct answer as well as the salient sentence in a multi-document context in order to learn cross-text relationships. 

\paragraph{Data Generation Process.} In order to facilitate the construction of a multi-document context, we propose three different modes, each one is responsible for uncovering information by using different contexts. For all the modes, we first generate a QA pair out of the most salient sentence in the held-out document.

\newlength{\textfloatsepsave} 
\setlength{\textfloatsepsave}{\textfloatsep}
\setlength{\textfloatsep}{0pt}
\begin{algorithm}[t]
\caption{Pre-training Data Generation}
\label{alg:overall}
\small{
\KwIn{
A text corpus of document clusters $\mathcal{C}=\{\mathcal{S}_1,...,\mathcal{S}_{|\mathcal{C}|}\}$, and a question-answer generator $\mathrm{\qasem{}(}\cdot)$.
}
\KwOut{The pre-training dataset $\mathcal{D}$.}
}
$\mathcal{D} \gets \emptyset$\;
\For{$n \gets 1$ to $|\mathcal{C}|$}
{
    \For{$k \gets 1$ to $|\mathcal{S}_n|$}
    {
        $s^*_k \gets \argmax\limits_{i}\text{CD-GSG-Score(}s_k^i)$\;
        $(q_k^*, a_k^*) \gets \mathrm{\qasem{}(}s_k^*)$\;
        $t_k^* = [a_k^*,s^*_k]$ \# target text\;
        $\mathcal{D} \gets \mathcal{D} \cup \left\{\left(\left[\mathcal{S}_n/D_k, q_k^*\right],t_k^*\right)\right\}$ \# (a)\;
        $\mathcal{D} \gets \mathcal{D} \cup \left\{\left(\left[\mathcal{S}_n/\left\{s_k^*\right\}, q_k^*\right],t_k^*\right)\right\}$ \# (b)\;
        $\mathcal{D} \gets \mathcal{D} \cup \left\{\left(\left[\mathcal\mathcal{S}_n/\left\{a_k^*\right\}, q_k^*\right],t_k^*\right)\right\}$ \# (c)\;
    }
}
Return $\mathcal{D}$\;
\end{algorithm}
\setlength{\textfloatsep}{\textfloatsepsave}

\begin{enumerate}[label=(\alph*),wide]
\item \textbf{Excluding the source document.} In this mode we disregard the held-out document $D_k$ from the context $\mathcal\mathcal{S}_n$ given to the model, i.e, $\mathcal{S}_n/D_k$. Hence, the model is tasked to predict the answer without having access to the source document at all, and is restricted to observe only the other documents in the set. Thus, this mode is considered as the most challenging one. 

\item \textbf{Masking the salient sentence.}  In this mode, the source salient sentence is masked, i.e, $\mathcal{S}_n/\left\{s_k^*\right\}$. The model has access to the surrounding context of the masked sentence in the held-out document, as well as the other documents in the set.

\item \textbf{Masking the answer.} In this mode, only the answer span within the salient sentence is masked, i.e, $\mathcal\mathcal{S}_n/\left\{a_k^*\right\}$. The model has access to the surrounding salient sentence, as well as all the documents in the set.
\end{enumerate}

As part of the new pre-training process of our novel multi-document model, we append the question after the context and instruct the model to generate an answer followed by its salient sentence, i.e., $output=\texttt{$\langle$answer$\rangle$, $\langle$sentence$\rangle$}$, inspired by \citet{Bohnet2022AttributedQA}. Generating the salient sentence introduces a copying mechanism (allows the model to also learn to copy information from the source directly) as well as allowing long-text generation, which is crucial for summarization downstream tasks~\cite{pegasus}, as well as outperforming a model which was pre-trained for generating the answer solely -- according to the ablations study, this setup yields the best performance results (\S\ref{subsec:ablation}). In the pre-training evaluation phase, the held-out set was split and the loss was measured separately for each mode of the data. As expected, we observed that the loss for (a) was significantly higher than those for the other modes, with (a)$\succ$(b)$\succ$(c) ranking highest. The procedure for generating the pre-training data is summarized in Algorithm~\ref{alg:overall} and Figure~\ref{figure:data}.

\paragraph{The resulted pre-training corpus.} We applied our procedure over the NewSHead corpus~\cite{newshead}, which consists of a set of related documents per instance. This is the exact same pre-training corpus used also by our main baseline \primera{}~\cite{xiao-etal-2022-primera} (See App.~\ref{app:data} for more details).

Using our data generation procedure, we produced 3,579,323 pre-training examples and 13,475 held-out examples, where on average, every 3.5 instances originated from the same cluster of related documents. In Table~\ref{tab:pretraining_stats}, we depict the comparison of pre-training corpora for related multi-document LMs compared to our \model{} pre-training data. 

\begin{table}[!tb]
\renewcommand{\arraystretch}{1.2}
      \centering
\footnotesize
\def\arraystretch{1.13}\setlength{\tabcolsep}{3pt} 
        \begin{tabular}[b]{@{}llcc@{}}

\toprule
    Model& Pretraining Dataset& \#clusters &\#instances \\  \toprule
\cdlm{}  ~\citeyearpar{caciularu-etal-2021-cdlm-cross}       & Multi-News~\citeyearpar{fabbri-etal-2019-multi} & 56K &   56K  \\ 
\primera{}~\citeyearpar{xiao-etal-2022-primera}  & NewSHead~\citeyearpar{newshead}   & \textbf{367K}  & 367K \\ 
\model{} (ours)  &  NewSHead~\citeyearpar{newshead}    & \textbf{367K} & \textbf{4.3M}\\
\toprule
\end{tabular}
\caption{Pre-training corpus statistics used by multi-document models. The reported numbers are the count of document clusters and the count of unique pre-training instances.}
\label{tab:pretraining_stats}
\end{table}
\section{Experimental Setup and Results}
\label{sec:eval}
This section presents experiments conducted to evaluate \model{},
as well as the the ablations and baselines we used. For the intrinsic evaluation we evaluated the models over multi-document QA tasks. For extrinsic evaluations we considered the multi-document abstractive summarization task.

\paragraph{Model Implementation Details} %
Following~\citet{xiao-etal-2022-primera}, we use the large-sized Longformer-Encoder-Decoder (LED) \cite{beltagy2020longformer} for our model initialization. 
The length limits of input and output are 4096 and 1024, respectively.\footnote{The tasks in this work consume inputs of up to 4k tokens.} Following the Huggingface implementation~\cite{wolf-etal-2020-transformers}, we set the sliding window size to $1024$ for local attention in the encoder part.

Similar to the \primera{} model \cite{xiao-etal-2022-primera}, when concatenating the documents and the question, we add a special document separator token (\texttt{<doc-sep>}) between the documents to signal to the model to be aware of the document boundaries. We also assign the global attention mode to these tokens which enables the model to share information across documents~\cite{caciularu-etal-2021-cdlm-cross}. For further hyperparameter and pre-training execution details, see App.~\ref{app:pretraining}.

\subsection{Multi-Document Question Answering}
\label{subsec:qa_eval}
Multi-document QA is the task of generating the correct answer, given a set of related multiple documents. For several multi-document QA benchmarks, models are often tasked to implicitly solve multiple sub-tasks or follow intermediate steps, such as comprehending the question, filtering out distracting documents in the context, and stitching pieces of information across the relevant documents~\cite{geva-etal-2021-whats,caciularu-etal-2022-long}.
Recall that \model{} was pre-trained over a automatically generated multi-document QA dataset. Hence, as a preliminary assessment, we first investigate \model{}'s performance over two multi-document QA benchmarks, HopotQA-distractor~\cite{yang-etal-2018-hotpotqa} and WikiHop~\cite{welbl-etal-2018-constructing} (see more details of the datasets in App.~\ref{app:qa}), and compare to other models that were pre-trained using underling un-masking objectives.

\paragraph{Fine-Tuning Format.} To follow our pre-training scheme, we append the question to the context and fine-tune the model to generate the correct answer. We use the Longformer Encoder-Decoder (LED)~\cite{beltagy2020longformer} and \primera{}~\cite{xiao-etal-2022-primera} as the baselines, for assesing the contribution of our pre-trainig format. Confirmed by \citet{beltagy2020longformer}, we found out that appending the \texttt{question:} and \texttt{context:} prefixes before the question and the context tokens, respectively, resulted in better performance.

\paragraph{Baselines.} We compare \model{} (447M parameters) against a set of strong long-context transformer baselines, including LED (447M parameters)~\cite{beltagy2020longformer}, \primera{} (447M parameters)~\cite{xiao-etal-2022-primera},\footnote{Note that our model used the exact same pre-training corpus and started from the same base model (LED) as \primera{}, which allows a direct comparison.} and LongT5-xl (3B parameters)\footnote{We report the largest, best performing model LongT5-xl from~\citet{guo-etal-2022-longt5}, supporting input lengths of up to 8k. Note that unlike \model{} the global attention pattern of this model is not optimal for multi-document contexts (see the paper for further details).}~\cite{guo-etal-2022-longt5} (see \S\ref{sec:related_work}).\footnote{While LED and LongT5 are not exclusively designed for multi-document tasks, they are strong and natural baselines to consider, as they process long input contexts, and report extensive evaluations on multi-document tasks.}

\begin{table}[t]
\centering
\footnotesize
\setlength{\tabcolsep}{9pt} \def\arraystretch{1.2}
        \begin{tabular}[b]{@{}clrr@{}}
            \toprule
            &Model            & F1 & EM \\ \midrule
            \multirow{3}{*}{\rotatebox[origin=c]{90}{HotpotQA}}
            &LED \cite{beltagy2020longformer}  & 65.8   & 50.6       \\ 
            & LongT5-xl~\cite{guo-etal-2022-longt5} & 66.1 & 50.9 \\
            &\primera{} \cite{xiao-etal-2022-primera} & 65.4 & 47.8 \\
            &\model{}& \textbf{67.1} & \textbf{52.7} \\
            \cmidrule(lr){2-4}
            \multirow{3}{*}{\rotatebox[origin=c]{90}{WikiHop}}
            &LED \cite{beltagy2020longformer}  & 65.6   & 62.4         \\ 
            & LongT5-xl~\cite{guo-etal-2022-longt5} & 67.7 & 63.6 \\
            &\primera{} \cite{xiao-etal-2022-primera} & 65.0 & 61.9 \\
            &\model{}& \textbf{69.3} & \textbf{65.2} \\

            \bottomrule
        \end{tabular}
\caption{HotpotQA-distractor and WikiHop results ($F_1$ and Exact Match) over the dev set.}
\label{tab:hotpotqa}
\end{table}

\paragraph{Results.} The results on multi-document QA are shown in Table \ref{tab:hotpotqa}. We adopted the F1 and Exact Match (EM) evaluation metrics corresponding to the original works. Our \model{} outperforms both \primera{}, LED, and LongT5, confirming that our pre-training data and input format are beneficial for both capturing cross-document relationships (\model{}$\succ$LED) as well as exploiting both context and question (\model{}$\succ$\primera{}).

\subsection{Multi-Document Summarization (MDS)}
\label{subsec:summ_eval}

This task aims at generating a summary for a given set of topically-related documents. Inherently, end-to-end MDS needs to implicitly address several subtasks including salience detection, redundancy removal, and text generation. Since dealing with multiple documents, MDS requires dealing with heterogeneous information and dispersed, while exhibiting substantial textual redundancy. We train and test \model{} with two challenging MDS benchmarks, each one dealing with a different domain: Multi-News~\cite{fabbri-etal-2019-multi}, which is concerned on summarizing related news articles, and Multi-XScience~\cite{lu-etal-2020-multi-xscience}, for scientific articles summarization (see more details of the datasets in App.~\ref{app:summ}). Under this setting, we are provided sets of documents (without any query), and therefore we simply encode the documents using \model{} without appending additional text. 

\paragraph{Baselines.} As in the previous experiment, we compare \model{} against LED, \primera{}, LongT5-xl. Following \citet{xiao-etal-2022-primera} we report the results of the state-of-the-art models from~\citet{pasunuru-etal-2021-efficiently} and \citet{lu-etal-2020-multi-xscience}, for Multi-News and Multi-XScience, respectively.

\paragraph{Results.} Tables \ref{tab:multinews} and \ref{tab:multixscience} present the evaluation results over the Multi-News and Multi-XScience datasets, respectively. Following previous MDS works, we report the \textsc{Rouge} R-1, -2, and -L scores, which are the standard MDS evaluation metrics (see App.~\ref{app:summ} for details).
For a fair comparison, we include the results of \primera{} as well as the results of the previous state-of-the-art methods (\citet{pasunuru-etal-2021-efficiently} and \citet{lu-etal-2020-multi-xscience}, for Multi-News and for Multi-XScience, respectively), and LED~\cite{beltagy2020longformer}. 
As shown in the results tables, \model{} exhibits the best performance across most of the examined models and benchmarks, especially on the Multi-News dataset, clearly demonstrating its consistent advantage. This excludes the results for Multi-XScience where \model{} slightly underperforms the prior work and LongT5. An explanation which \citet{xiao-etal-2022-primera} points refers to the fact that the clusters in Multi-XScience have less overlapping information compared to the corpus we used, attributed to the use of abstracts as the input documents in Multi-XScience. In addition, LongT5 advantage over \model{} is attributed to significantly larger number of parameters of LongT5-xl.

\begin{table}[t]
\centering
\footnotesize
\setlength{\tabcolsep}{9pt} 
\renewcommand{\arraystretch}{1.2}
        \begin{tabular}[b]{@{}lrrr@{}}
            \toprule

            Model            & R-1 & R-2 & R-L  \\ \midrule
           
            \citet{pasunuru-etal-2021-efficiently}& 49.2 & 19.6 & 24.5\\
            LED~\cite{beltagy2020longformer} & 47.4 & 20.7 & 23.7 \\
            LongT5-xl~\cite{guo-etal-2022-longt5} & 47.4 & 20.7 & 23.7 \\
            \primera{}~\cite{xiao-etal-2022-primera} & 49.9 & 21.1 & 25.9 \\
            \model{} & \textbf{50.9} & \textbf{23.1} & \textbf{27.2} \\
            \bottomrule
        \end{tabular}
\caption{\textsc{Rouge} (-1,-2,-L) results for the test set of the Multi-News dataset.} 
\label{tab:multinews}
\end{table}

\begin{table}[t]
\centering
\footnotesize
\setlength{\tabcolsep}{9pt} 
\renewcommand{\arraystretch}{1.2}
        \begin{tabular}[b]{@{}lrrr@{}}
            \toprule

            Model            & R-1 & R-2 & R-L  \\ \midrule
           
            \citet{lu-etal-2020-multi-xscience}& \textbf{33.9} & 6.8 & 18.2 \\
            LED~\cite{beltagy2020longformer} & 31.0 & 6.9 & 17.4 \\
            LongT5-xl~\cite{guo-etal-2022-longt5} & 33.7 & \textbf{8.1} & \textbf{19.4} \\
            \primera{}~\cite{xiao-etal-2022-primera} & 31.9 & 7.4 & 18.0 \\
            \model{} & 33.5 & 7.6 & 19.1\\
            \bottomrule
        \end{tabular}
\caption{\textsc{Rouge} (-1,-2,-L) results for the test set of the Multi-XScience dataset.} 
\label{tab:multixscience}
\vspace{-3mm}
\end{table}

\subsection{Query-Focused Multi-Document Abstractive Summarization}
\label{subsec:qsumm_eval}

The task of Query-focused Multi-Document Summarization (QMDS) aims at generating a summary from a set of documents, that answers a specific given query. Unlike MDS, QMDS tries to solve more realistic query-based scenarios, since it suggests summarizing only predefined salient information of interest that best answers the query. Since we proposed pre-trainng under the multi-document question answering setup, we posit that \model{} might be effective for QMDS.

We consider the datasets constructed by \citet{pasunuru2021data}, \textsc{QmdsCnn} and \textsc{QmdsIr} (see more details of the datasets in App.~\ref{app:qfsumm}) as well as their strong baseline, and include also the results of \primera{} and LED.

\paragraph{Baselines.} Similar to the previous experiments, we compare \model{} against LED, \primera{}, LongT5-xl. In addition, we consider also the baseline from \citet{pasunuru2021data}.

\patchcmd{\footnotemark}{\stepcounter{footnote}}{\refstepcounter{footnote}}{}{}
\begin{table}[t]
\centering
\footnotesize
\setlength{\tabcolsep}{9pt} 
\renewcommand{\arraystretch}{1.2}
        \begin{tabular}[b]{@{}lrrr@{}}
            \toprule

            Model            & R-1 & R-2 & R-L  \\ \midrule
           
            \citet{pasunuru2021data}\footnotemark\label{note1}& 37.9 & 16.4 & 35.2\\
            LED~\cite{beltagy2020longformer} & 32.3 & 14.3 & 30.9 \\
            LongT5-xl~\cite{guo-etal-2022-longt5} & 35.5 & 15.9 & 34.3 \\
            \primera{}~\cite{xiao-etal-2022-primera} & 36.1 & 16.2 & 35.7 \\
            \model{} & \textbf{38.8} & \textbf{18.3} & \textbf{37.2} \\
            \bottomrule
        \end{tabular}
\caption{\textsc{Rouge} (-1,-2,-L) results for the test set of the \textsc{QmdsCnn} dataset.} 
\label{tab:qmdscnn}
\end{table}

\begin{table}[t]
\centering
\footnotesize
\setlength{\tabcolsep}{9pt} 
\renewcommand{\arraystretch}{1.2}
        \begin{tabular}[b]{@{}lrrr@{}}
            \toprule
            Model            & R-1 & R-2 & R-L  \\ \midrule
            \citet{pasunuru2021data}\textsuperscript{\ref{note1}}& 45.5 & 23.4 & 41.2 \\
            LED~\cite{beltagy2020longformer} & 43.2 & 21.3 & 40.5 \\
            LongT5-xl~\cite{guo-etal-2022-longt5} & 44.4 & 22.3 & 40.0 \\
            \primera{}~\cite{xiao-etal-2022-primera} & 45.7 & 23.6 & 40.9  \\
            \model{} & \textbf{47.6} & \textbf{25.1} & \textbf{42.4}\\
            \bottomrule
        \end{tabular}
\caption{\textsc{Rouge} (-1,-2,-L) results for the test set of the \textsc{QmdsIr} dataset.} 
\label{tab:qmdsir}
\vspace{-3mm}
\end{table}

\paragraph{Results.} Tables \ref{tab:qmdscnn} and \ref{tab:qmdsir} present the evaluation results over the \textsc{QmdsCnn} and \textsc{QmdsIr} datasets, respectively. Following MDS tasks and \citet{pasunuru2021data}, we report the \textsc{Rouge} R-1, -2, and -L scores, which are the standard MDS evaluation metrics (see App.~\ref{app:qfsumm} for details). As shown in the tables, \model{} exhibits the best performance across most of the examined models and benchmarks, clearly demonstrating its consistent advantage over the baselines. 

\footnotetext{We report the results of the best ablated model from \citet{pasunuru2021data}.}

\subsection{Ablation Study}
\label{subsec:ablation}

\paragraph{Data Generation.} We next turn to a broad ablation study, for assessing our configuration and design choices across our suggested pipeline. First, we show the advantage of combining the three proposed data modes, rather than using a subset of them. We evaluate all the resulted models by fine-tuning them over HopotQA-distractor (\S \ref{subsec:qa_eval}), Multi-XScience (\S \ref{subsec:summ_eval}), and \textsc{QmdsIR} (\S \ref{subsec:qsumm_eval}). For HopotQA-distractor we report the Exact Match (EM) score, and for the summarization tasks we report the \textsc{Rouge}-1 (R-1) score.

\paragraph{Baselines.} We pre-train \model{} for 100k steps, for using every subset of the set of the set (superset) of modes $\{(a),(b),(c)\}$ (all its possible combinations) of the generated pre-training data modes presented in \S\ref{sec:data_creation}. Note that our  \model{} model is referred to as using all the modes, i.e., $(a)+(b)+(c)$.

\begin{figure}[t]
    \centering
    \includegraphics[scale=.56]{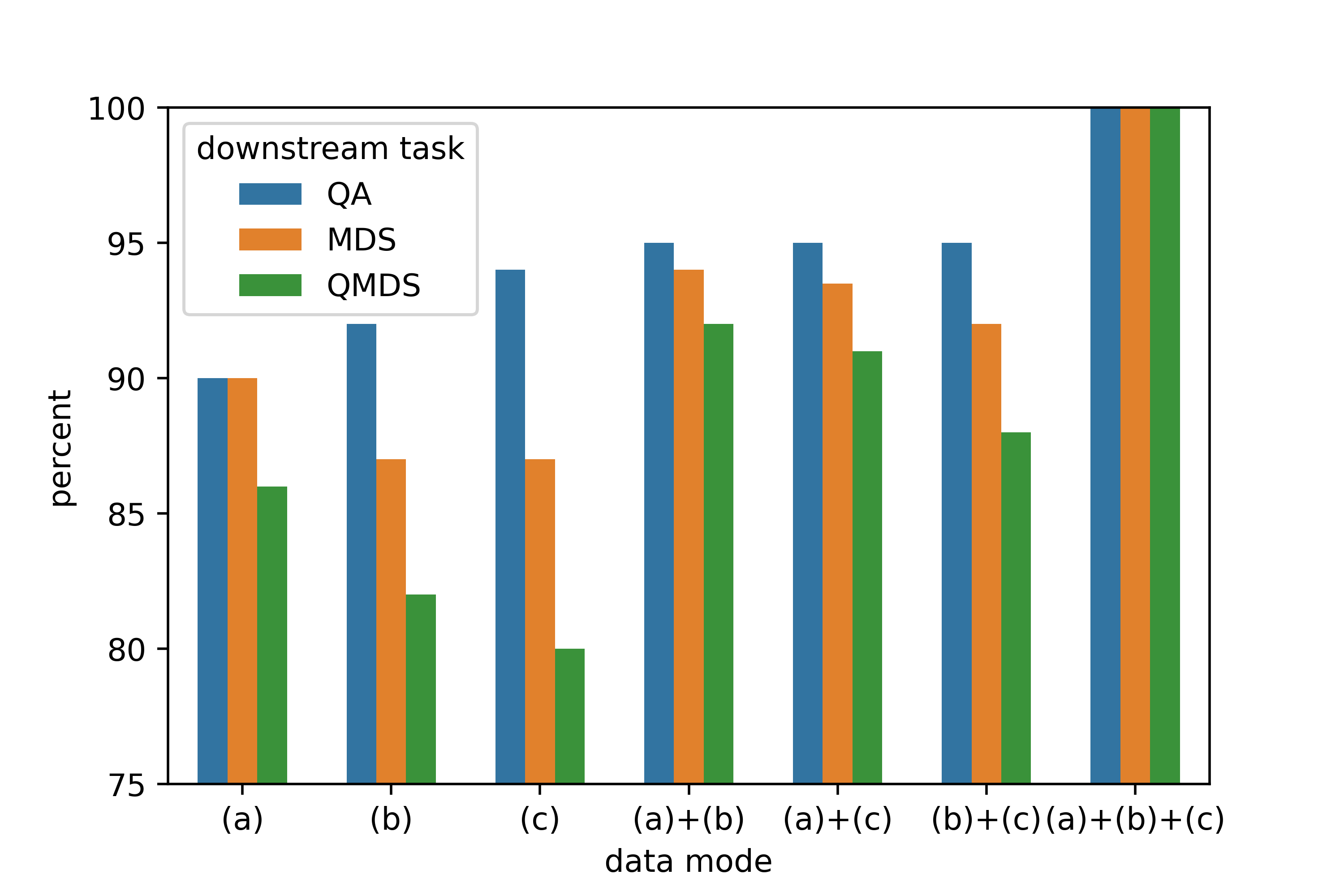}
    \caption{Ablation results over the validation sets of the  HotpotQA-distractor (QA) Multi-XScience (MDS), and \textsc{QmdsIR} (QMDS) datasets. We report the percentage of the perforamnce relatively to the top scoring model. For QA we used the EM score, and for MDS and QMDS we used the \textsc{Rouge}-1 score.}
    \label{figure:data_ablation}
    \vspace{-3mm}
\end{figure}

\paragraph{Results.} Figure \ref{figure:data_ablation} shows the ablation results. In all tasks, pre-training using all modes yields the best results. Among all modes, mode (c) appears to be the most effective for QA, since this is an extractive QA task, and mode (c) provides data in this format. Mode (a) excels at the summarization tasks, attributed to their abstractive nature as well as the requirement of all the documents for generating appropriate summaries.

\paragraph{Input Format}
We repeat the previous experiment and ablate the pre-training input format according to the multiple  different formats, and compare to the model pre-training format described in \S\ref{sec:data_creation} (with the same pre-training data): \textit{without questions}, \textit{with random question}, \textit{with random context document}, \textit{with prefixes}, \textit{placing the question before the context}, \textit{with question filtering}, and \textit{without generating the salient sentence}. Additionally, we assess the choice of \qasem{} as our question-answer generation module by using the generators from \citet{jia-etal-2022-question} and \citet{khashabi2022unifiedqa}. Finally, we also include the results of \primera{}, which was further pre-trained for additional 300k steps (fine-tuning LED for 400k steps in total), for a fair comparison to \model{} ablated models. See full details regarding all the ablations in App.~\ref{app:ablations}.

\begin{table}[!tb]
    \newcommand{\colindent}{\;}
    \centering
    \resizebox{0.49\textwidth}{!}{
    \begin{tabular}{@{}lcccc@{}}
    \toprule
    \phantom{fwidsvhckzxjvchndfzxvvgdaczfc} & QA & MDS & QMDS\\
    \midrule
     \textit{without questions}  & 60.3  & 32.8 & 44.7\\
    \textit{with random questions} & 61.1 & 32.1 & 44.2\\
    \textit{with random context documents} & 61.0  & 31.5 & 43.9 \\
    \textit{with prefixes} &\textbf{67.3} &32.6 &46.2 \\
    \textit{placing the question before the context} & 66.7 & \textbf{33.4} & 46.3 \\
    \textit{with question filtering} & 65.2 & 30.9 & 41.1 \\
    \textit{without generating the salient sentence} & 66.6 & 30.5 & 42.8 \\
        \midrule
    Using \citet{jia-etal-2022-question} as the QA generator & 66.6 & 33.2 & 45.9 \\
    Using \citet{khashabi2022unifiedqa} as the QA generator & 66.8 & 33.3 & 45.1 \\
        \midrule
    \primera{} \cite{xiao-etal-2022-primera} 400k steps checkpoint & 65.9 & 32.1 & 45.7 \\
    
            \midrule

    \colindent \model{}  &67.1 &\textbf{33.5} & \textbf{47.6} \\
    \bottomrule
    \end{tabular}}
    \caption{Ablation study results.}
    \label{tab:ablations}
    \vspace{-3mm}
\end{table}

\paragraph{Results.} Overall, our \model{} model outperforms the ablation models on most of the tasks, which a significant margin. 

Pre-training the model without any questions during or using random questions, negatively impacts the results of downstream tasks. An important function of the question is to facilitate the model's ability to generate the appropriate answer and the source sentence. This aligns with the findings from \citet{caciularu-etal-2021-cdlm-cross}, who showed that pre-training with random documents rather than related ones is sub-optimal.

The use of question and context prefixes for positioning input appears to be helpful for QA, but is inferior when applied to summarization tasks due to its unique format, which is well suited for QA but seems to generalize harder for other setups. When the question is placed before the context, performance slightly decreases over query-based tasks, while maintaining the same results for summarization (where the question location is irrelevant). 

Using question filtering is found to harm the downstream results of \model{}, in accordance to other QA-based pre-training prior works~\cite{jia-etal-2022-question}.

Pre-training without generating the attributed source sentence introduces a significant flow to the model, particularly for the summarization downstream tasks. As mentioned before, generating longer sequences, as well as teaching the model to copy text, is beneficial for summarization tasks. 

Applying a different question generator rather then \qasem{} yields inferior results overall, since the other generators produce open-ended questions and answers which are more prone to errors, while \qasem{} utilizes an existing span in the context as the answer. In addition, \qasem{} generated local questions, which allows \model{} to focus on the fine-grained details, and not only the coarse-grained information in the multi-document context.

When \primera{} is pre-trained with 400k steps (to match \model{}'s number of further pre-training steps), it underperforms \model{} and even fails to add any significant improvements over its 100K checkpoint, possibly due to the small amount of pre-training data it contains.

\begin{table}[t]
\centering
\footnotesize
\setlength{\tabcolsep}{9pt} 
\renewcommand{\arraystretch}{1.2}
        \begin{tabular}[b]{@{}lrrr@{}}
            \toprule
            Model            & R-1 & R-2 & R-L  \\ \midrule
            \primera{} & 45.0 & 16.7 & 22.6 \\
            GPT-3.5 & 36.4 & 10.8 &  18.7\\
            GPT-4 & 34.7& 10.7 & 18.8\\
            GPT-4 8k & 34.9 & 10.9 & 18.9 \\
            \model{} & \textbf{45.3} & \textbf{17.4} & \textbf{23.7} \\
            \bottomrule
        \end{tabular}
\caption{\textsc{Rouge} (-1,-2,-L) results on a subset of Multi-News. GPT models are accessed through the OpenAI public API and are applied in zero-shot mode.}
\label{tab:gpt}
\vspace{-3mm}
\end{table}
\begin{table}[t]
\centering
\footnotesize
\setlength{\tabcolsep}{7pt} 
\renewcommand{\arraystretch}{1.2}
        \begin{tabular}[b]{@{}lcccc@{}}
            \toprule

            Model            & Cont. & Read. & Gram. & Non-red. \\ \midrule
            \primera{} & $\uparrow$53.3\% & $\uparrow$63.3\% & $\uparrow$56.7\% & $\uparrow$53.3\% \\
            GPT-3.5 & $\uparrow$70.0\% & $\downarrow$33.3\% & $\downarrow$30.0\% & $\uparrow$70.0\% \\
            GPT-4 8k &  $\uparrow$73.3\% & $\downarrow$40.0\% & $\downarrow$36.6\% & $\uparrow$83.3\% \\
            \bottomrule
        \end{tabular}
\caption{Comparison of the first 30 summaries of the Multi-News sample between \model{} and the baselines. Under each of the four evaluation criteria, the cells in a row indicate the percentage of cases where our system was preferred over the GPT baseline.}
\label{tab:gpt_human}
\vspace{-4mm}
\end{table}

\subsection{Comparison with Large Language Models}
In order to get insights into how \model{} compares with state-of-the-art Generalist Large Language Models (LLMs), we provide a small comparison with two capable models, GPT-3.5 turbo~\cite{NEURIPS2022_b1efde53} and GPT-4\footnote{Versions available and accessed on 5/15/23. It should be noted that data contamination with Multi-News is highly likely~\cite{jacovi2023stop}.}~\cite{OpenAI2023GPT4TR} (including the 8k input length version) evaluated on the zero-shot setting.

For a fair comparison, we used the same context window size of 4K tokens for all models (and up to 8k for GPT-4 8k). Due to the fact that multi-document tasks involve processing long sequences, the cost of API calls is significant for a comprehensive evaluation across all datasets. Therefore, we only evaluate on a sample of 200 instances from the multi-news dataset (see prompting details in App. \ref{sec:prompt}). Table \ref{tab:gpt} depicts the results. We observe that \model{} significantly outperforms both GPT-3.5 and GPT-4 models, though the performance of GPT-4 and GPT-3.5 is comparable. We leave more comprehensive comparisons with LLMs to future work.

We further assessed \model{} through manual comparison against \primera{}, GPT-3.5, and GPT-4 8k. NLP graduate students were shown summaries for a given topic from the three systems and \model{} in arbitrary order, along with a corresponding reference summary. Following \cite{ernst-etal-2022-proposition}, participants were asked to rank the systems based on Content (overlap with the reference), Readability (the readability of a summary), Grammaticality (avoiding grammar errors), and Non-Redundancy (avoiding  repetitions), and we extract the pairwise results out of the rankings (see \cite{ernst-etal-2022-proposition} for further details). In App.~\ref{sec:summ_examples}, we provide several examples to system summaries and their corresponding reference summaries.

The results of this study are presented in Table \ref{tab:gpt_human}. Under each evaluation criterion, it indicates the percentage of cases where \model{} was preferred over both baselines. \model{} was favored in all cases except for grammatical errors and readability (which corresponds to the Reinforcement Learning from Human Feedback phase of the GPT models).
\section{Conclusions}
\label{sec:conclusions}
In this work, we present a novel pre-training scheme for multi-document tasks. First, our approach suggests to augment the existing multi-document pre-training objectives into a cross-document question answering task. Second, we generate high-quality large-scale QA pre-training data using a controlled generation approach, in which each QA pair originates from a salient sentence in one of the documents in the set.

During pre-training, we task the the Longformer Encoder-Decoder (LED) model to generate the answer and the salient sentence on the basis of the remaining context. This objective encourages the LED model to elicit cross-document relationships, and stitch pieces of information across the input documents, which are relevant for performing multi-document tasks. The resulted model \model{} shows significant performance improvements compared to prior models under extensive experimentation over multiple challenging multi-document summarization and QA datasets. 

Future work can extend the ideas in this work for equipping decoder-only large LMs with cross-document modeling using our proposed method, also in the setup of in-context learning and prompt tuning. We foresee that our method should be significant specifically for retrieval-augmented language modeling setups~\cite{izacard2022few}, where there is a use of related documents as an outsourced external non-parametric knowledge source. Finally, the use of a single document in order to trigger cross-document relationships, as firstly introduced in this work, might be further investigated.

\section*{Acknowledgements}
The work described herein was supported by the PBC fellowship for outstanding PhD candidates in data science, in part by grants from the Israel Science Foundation grant 2827/21, and by a grant from the Israel Ministry of Science and Technology.   

\section*{Limitations}
While our work tries to focus around reasoning over both fine- and coarse-grained cross-document relationships, \model{}, the resulted pre-trained model, might still suffer from factual consistency errors while generating information given a query, and there is no guarantee that it will always generate factual and reasonable content without any further fine-tuning.

The \qasem{} question generation model that we used may also have been a source of these problems. There is a possibility that \qasem{} produces inadequate questions that could harm the pre-training process of the model. An attempt was made to filter out noise using a question model, but the results were inferior to non-filtering. Consequently, if the model is not fine-tuned, inconsistency (hallucinations) may occur more frequently.

In addition, by using the Newshead corpus as the pre-training data source, we assume that it is comprised of high quality documents. We also take into account the fact that Newshead is limited to documents in the news domain, while some of the benchmarks used for evaluating \model{} include another topics of interest. Future work may further assess the quality of the documents, such as checking for duplications or wrong statements, and diversify the corpus domains. This is crucial for productizing models like \model{} in interactive multi-text applications (chatbots) and semantic search applications which are gaining attraction nowadays~\cite{hirsch-etal-2021-ifacetsum,eirew-etal-2022-cross}.

Finally, the resulted model \model{} was pre-trained on sets of related documents, by answering questions that matched their content. As in an out-of-domain scenario, \model{}'s use over sets of documents that are not related, or over single documents, might be unexpected. Such settings may be the subject of another research direction in the future.

\section*{Ethics Statement}
Despite the limited risk associated with our work, similar to existing state-of-the-art generation language models, there is no guarantee that \model{}, our model, will always generate factual information. The model should therefore be used with caution in a practical environment and be carefully tested before deployment. It is possible, for example, that frequent anecdotal events in the pre-training dataset are generated in an unexpected manner.

\bibliography{all,anthology}

\begin{thebibliography}{57}
\expandafter\ifx\csname natexlab\endcsname\relax\def\natexlab#1{#1}\fi

\bibitem[{Ainslie et~al.(2020)Ainslie, Ontanon, Alberti, Cvicek, Fisher, Pham,
  Ravula, Sanghai, Wang, and Yang}]{ainslie-etal-2020-etc}
Joshua Ainslie, Santiago Ontanon, Chris Alberti, Vaclav Cvicek, Zachary Fisher,
  Philip Pham, Anirudh Ravula, Sumit Sanghai, Qifan Wang, and Li~Yang. 2020.
\newblock \href {https://doi.org/10.18653/v1/2020.emnlp-main.19} {{ETC}:
  Encoding long and structured inputs in transformers}.
\newblock In \emph{Proceedings of the 2020 Conference on Empirical Methods in
  Natural Language Processing (EMNLP)}, pages 268--284, Online. Association for
  Computational Linguistics.

\bibitem[{Alberti et~al.(2019)Alberti, Andor, Pitler, Devlin, and
  Collins}]{alberti-etal-2019-synthetic}
Chris Alberti, Daniel Andor, Emily Pitler, Jacob Devlin, and Michael Collins.
  2019.
\newblock \href {https://doi.org/10.18653/v1/P19-1620} {Synthetic {QA} corpora
  generation with roundtrip consistency}.
\newblock In \emph{Proceedings of the 57th Annual Meeting of the Association
  for Computational Linguistics}, pages 6168--6173, Florence, Italy.
  Association for Computational Linguistics.

\bibitem[{Beltagy et~al.(2020)Beltagy, Peters, and
  Cohan}]{beltagy2020longformer}
Iz~Beltagy, Matthew~E Peters, and Arman Cohan. 2020.
\newblock Longformer: The long-document transformer.
\newblock \emph{arXiv preprint arXiv:2004.05150}.

\bibitem[{Bohnet et~al.(2022)Bohnet, Tran, Verga, Aharoni, Andor, Soares,
  Eisenstein, Ganchev, Herzig, Hui, Kwiatkowski, Ma, Ni, Schuster, Cohen,
  Collins, Das, Metzler, Petrov, and Webster}]{Bohnet2022AttributedQA}
Bernd Bohnet, Vinh~Q. Tran, Pat Verga, Roee Aharoni, Daniel Andor,
  Livio~Baldini Soares, Jacob Eisenstein, Kuzman Ganchev, Jonathan Herzig, Kai
  Hui, Tom Kwiatkowski, Ji~Ma, Jianmo Ni, Tal Schuster, William~W. Cohen,
  Michael Collins, Dipanjan Das, Donald Metzler, Slav Petrov, and Kellie
  Webster. 2022.
\newblock Attributed question answering: Evaluation and modeling for attributed
  large language models.
\newblock \emph{arXiv preprint arXiv:2212.08037}, 4.

\bibitem[{Caciularu et~al.(2021)Caciularu, Cohan, Beltagy, Peters, Cattan, and
  Dagan}]{caciularu-etal-2021-cdlm-cross}
Avi Caciularu, Arman Cohan, Iz~Beltagy, Matthew Peters, Arie Cattan, and Ido
  Dagan. 2021.
\newblock \href {https://doi.org/10.18653/v1/2021.findings-emnlp.225} {{CDLM}:
  Cross-document language modeling}.
\newblock In \emph{Findings of the Association for Computational Linguistics:
  EMNLP 2021}, pages 2648--2662, Punta Cana, Dominican Republic. Association
  for Computational Linguistics.

\bibitem[{Caciularu et~al.(2022)Caciularu, Dagan, Goldberger, and
  Cohan}]{caciularu-etal-2022-long}
Avi Caciularu, Ido Dagan, Jacob Goldberger, and Arman Cohan. 2022.
\newblock \href {https://doi.org/10.18653/v1/2022.naacl-main.207} {Long context
  question answering via supervised contrastive learning}.
\newblock In \emph{Proceedings of the 2022 Conference of the North American
  Chapter of the Association for Computational Linguistics: Human Language
  Technologies}, pages 2872--2879, Seattle, United States. Association for
  Computational Linguistics.

\bibitem[{Eirew et~al.(2022)Eirew, Caciularu, and
  Dagan}]{eirew-etal-2022-cross}
Alon Eirew, Avi Caciularu, and Ido Dagan. 2022.
\newblock \href {https://aclanthology.org/2022.emnlp-main.58} {Cross-document
  event coreference search: Task, dataset and modeling}.
\newblock In \emph{Proceedings of the 2022 Conference on Empirical Methods in
  Natural Language Processing}, pages 900--913, Abu Dhabi, United Arab
  Emirates. Association for Computational Linguistics.

\bibitem[{Ernst et~al.(2022)Ernst, Caciularu, Shapira, Pasunuru, Bansal,
  Goldberger, and Dagan}]{ernst-etal-2022-proposition}
Ori Ernst, Avi Caciularu, Ori Shapira, Ramakanth Pasunuru, Mohit Bansal, Jacob
  Goldberger, and Ido Dagan. 2022.
\newblock \href {https://doi.org/10.18653/v1/2022.naacl-main.128}
  {Proposition-level clustering for multi-document summarization}.
\newblock In \emph{Proceedings of the 2022 Conference of the North American
  Chapter of the Association for Computational Linguistics: Human Language
  Technologies}, pages 1765--1779, Seattle, United States. Association for
  Computational Linguistics.

\bibitem[{Fabbri et~al.(2019)Fabbri, Li, She, Li, and
  Radev}]{fabbri-etal-2019-multi}
Alexander Fabbri, Irene Li, Tianwei She, Suyi Li, and Dragomir Radev. 2019.
\newblock \href {https://doi.org/10.18653/v1/P19-1102} {Multi-news: A
  large-scale multi-document summarization dataset and abstractive hierarchical
  model}.
\newblock In \emph{Proceedings of the 57th Annual Meeting of the Association
  for Computational Linguistics}, pages 1074--1084, Florence, Italy.
  Association for Computational Linguistics.

\bibitem[{Fang et~al.(2020)Fang, Wang, Gan, Sun, Liu, and
  Zhu}]{fang2020accelerating}
Yuwei Fang, Shuohang Wang, Zhe Gan, Siqi Sun, Jingjing Liu, and Chenguang Zhu.
  2020.
\newblock Accelerating real-time question answering via question generation.
\newblock \emph{arXiv preprint arXiv:2009.05167}.

\bibitem[{Fisch et~al.(2019)Fisch, Talmor, Jia, Seo, Choi, and
  Chen}]{fisch-etal-2019-mrqa}
Adam Fisch, Alon Talmor, Robin Jia, Minjoon Seo, Eunsol Choi, and Danqi Chen.
  2019.
\newblock \href {https://doi.org/10.18653/v1/D19-5801} {{MRQA} 2019 shared
  task: Evaluating generalization in reading comprehension}.
\newblock In \emph{Proceedings of the 2nd Workshop on Machine Reading for
  Question Answering}, pages 1--13, Hong Kong, China. Association for
  Computational Linguistics.

\bibitem[{FitzGerald et~al.(2018)FitzGerald, Michael, He, and
  Zettlemoyer}]{fitzgerald-etal-2018-large}
Nicholas FitzGerald, Julian Michael, Luheng He, and Luke Zettlemoyer. 2018.
\newblock \href {https://doi.org/10.18653/v1/P18-1191} {Large-scale {QA}-{SRL}
  parsing}.
\newblock In \emph{Proceedings of the 56th Annual Meeting of the Association
  for Computational Linguistics (Volume 1: Long Papers)}, pages 2051--2060,
  Melbourne, Australia. Association for Computational Linguistics.

\bibitem[{Geva et~al.(2021)Geva, Katz, Ben-Arie, and
  Berant}]{geva-etal-2021-whats}
Mor Geva, Uri Katz, Aviv Ben-Arie, and Jonathan Berant. 2021.
\newblock \href {https://doi.org/10.18653/v1/2021.emnlp-main.646} {{W}hat{'}s
  in your head? {E}mergent behaviour in multi-task transformer models}.
\newblock In \emph{Proceedings of the 2021 Conference on Empirical Methods in
  Natural Language Processing}, pages 8201--8215, Online and Punta Cana,
  Dominican Republic. Association for Computational Linguistics.

\bibitem[{Ginzburg et~al.(2021)Ginzburg, Malkiel, Barkan, Caciularu, and
  Koenigstein}]{ginzburg-etal-2021-self}
Dvir Ginzburg, Itzik Malkiel, Oren Barkan, Avi Caciularu, and Noam Koenigstein.
  2021.
\newblock \href {https://doi.org/10.18653/v1/2021.findings-acl.272}
  {Self-supervised document similarity ranking via contextualized language
  models and hierarchical inference}.
\newblock In \emph{Findings of the Association for Computational Linguistics:
  ACL-IJCNLP 2021}, pages 3088--3098, Online. Association for Computational
  Linguistics.

\bibitem[{Gu et~al.(2020)Gu, Mao, Han, Liu, Wu, Yu, Finnie, Yu, Zhai, and
  Zukoski}]{newshead}
Xiaotao Gu, Yuning Mao, Jiawei Han, Jialu Liu, You Wu, Cong Yu, Daniel Finnie,
  Hongkun Yu, Jiaqi Zhai, and Nicholas Zukoski. 2020.
\newblock Generating representative headlines for news stories.
\newblock In \emph{Proceedings of The World Wide Web Conference (WWW)}.

\bibitem[{Guo et~al.(2022)Guo, Ainslie, Uthus, Ontanon, Ni, Sung, and
  Yang}]{guo-etal-2022-longt5}
Mandy Guo, Joshua Ainslie, David Uthus, Santiago Ontanon, Jianmo Ni, Yun-Hsuan
  Sung, and Yinfei Yang. 2022.
\newblock \href {https://doi.org/10.18653/v1/2022.findings-naacl.55}
  {{L}ong{T}5: {E}fficient text-to-text transformer for long sequences}.
\newblock In \emph{Findings of the Association for Computational Linguistics:
  NAACL 2022}, pages 724--736, Seattle, United States. Association for
  Computational Linguistics.

\bibitem[{He et~al.(2020)He, Ning, and Roth}]{he-etal-2020-quase}
Hangfeng He, Qiang Ning, and Dan Roth. 2020.
\newblock \href {https://doi.org/10.18653/v1/2020.acl-main.772} {{Q}u{ASE}:
  Question-answer driven sentence encoding}.
\newblock In \emph{Proceedings of the 58th Annual Meeting of the Association
  for Computational Linguistics}, pages 8743--8758, Online. Association for
  Computational Linguistics.

\bibitem[{He et~al.(2015)He, Lewis, and Zettlemoyer}]{he-etal-2015-question}
Luheng He, Mike Lewis, and Luke Zettlemoyer. 2015.
\newblock \href {https://doi.org/10.18653/v1/D15-1076} {Question-answer driven
  semantic role labeling: Using natural language to annotate natural language}.
\newblock In \emph{Proceedings of the 2015 Conference on Empirical Methods in
  Natural Language Processing}, pages 643--653, Lisbon, Portugal. Association
  for Computational Linguistics.

\bibitem[{Hermann et~al.(2015)Hermann, Kocisky, Grefenstette, Espeholt, Kay,
  Suleyman, and Blunsom}]{NIPS2015_afdec700}
Karl~Moritz Hermann, Tomas Kocisky, Edward Grefenstette, Lasse Espeholt, Will
  Kay, Mustafa Suleyman, and Phil Blunsom. 2015.
\newblock Teaching machines to read and comprehend.
\newblock In \emph{Advances in Neural Information Processing Systems (NIPS)}.

\bibitem[{Hirsch et~al.(2021)Hirsch, Eirew, Shapira, Caciularu, Cattan, Ernst,
  Pasunuru, Ronen, Bansal, and Dagan}]{hirsch-etal-2021-ifacetsum}
Eran Hirsch, Alon Eirew, Ori Shapira, Avi Caciularu, Arie Cattan, Ori Ernst,
  Ramakanth Pasunuru, Hadar Ronen, Mohit Bansal, and Ido Dagan. 2021.
\newblock \href {https://doi.org/10.18653/v1/2021.emnlp-demo.33}
  {i{F}acet{S}um: Coreference-based interactive faceted summarization for
  multi-document exploration}.
\newblock In \emph{Proceedings of the 2021 Conference on Empirical Methods in
  Natural Language Processing: System Demonstrations}, pages 283--297, Online
  and Punta Cana, Dominican Republic. Association for Computational
  Linguistics.

\bibitem[{Hoffmann et~al.(2022)Hoffmann, Borgeaud, Mensch, Buchatskaya, Cai,
  Rutherford, Casas, Hendricks, Welbl, Clark et~al.}]{hoffmann2022training}
Jordan Hoffmann, Sebastian Borgeaud, Arthur Mensch, Elena Buchatskaya, Trevor
  Cai, Eliza Rutherford, Diego de~Las Casas, Lisa~Anne Hendricks, Johannes
  Welbl, Aidan Clark, et~al. 2022.
\newblock Training compute-optimal large language models.
\newblock \emph{arXiv preprint arXiv:2203.15556}.

\bibitem[{Huber et~al.(2022)Huber, Aghajanyan, Oguz, Okhonko, Yih, Gupta, and
  Chen}]{huber-etal-2022-ccqa}
Patrick Huber, Armen Aghajanyan, Barlas Oguz, Dmytro Okhonko, Scott Yih, Sonal
  Gupta, and Xilun Chen. 2022.
\newblock \href {https://doi.org/10.18653/v1/2022.findings-naacl.184} {{CCQA}:
  A new web-scale question answering dataset for model pre-training}.
\newblock In \emph{Findings of the Association for Computational Linguistics:
  NAACL 2022}, pages 2402--2420, Seattle, United States. Association for
  Computational Linguistics.

\bibitem[{Izacard et~al.(2022)Izacard, Lewis, Lomeli, Hosseini, Petroni,
  Schick, Dwivedi-Yu, Joulin, Riedel, and Grave}]{izacard2022few}
Gautier Izacard, Patrick Lewis, Maria Lomeli, Lucas Hosseini, Fabio Petroni,
  Timo Schick, Jane Dwivedi-Yu, Armand Joulin, Sebastian Riedel, and Edouard
  Grave. 2022.
\newblock Few-shot learning with retrieval augmented language models.
\newblock \emph{arXiv preprint arXiv:2208.03299}.

\bibitem[{Jacovi et~al.(2023)Jacovi, Caciularu, Goldman, and
  Goldberg}]{jacovi2023stop}
Alon Jacovi, Avi Caciularu, Omer Goldman, and Yoav Goldberg. 2023.
\newblock Stop uploading test data in plain text: Practical strategies for
  mitigating data contamination by evaluation benchmarks.
\newblock \emph{arXiv preprint arXiv:2305.10160}.

\bibitem[{Jia et~al.(2022)Jia, Lewis, and Zettlemoyer}]{jia-etal-2022-question}
Robin Jia, Mike Lewis, and Luke Zettlemoyer. 2022.
\newblock \href {https://doi.org/10.18653/v1/2022.findings-acl.59} {Question
  answering infused pre-training of general-purpose contextualized
  representations}.
\newblock In \emph{Findings of the Association for Computational Linguistics:
  ACL 2022}, pages 711--728, Dublin, Ireland. Association for Computational
  Linguistics.

\bibitem[{Khashabi et~al.(2022)Khashabi, Kordi, and
  Hajishirzi}]{khashabi2022unifiedqa}
Daniel Khashabi, Yeganeh Kordi, and Hannaneh Hajishirzi. 2022.
\newblock Unifiedqa-v2: Stronger generalization via broader cross-format
  training.
\newblock \emph{arXiv preprint arXiv:2202.12359}.

\bibitem[{Khashabi et~al.(2020)Khashabi, Min, Khot, Sabharwal, Tafjord, Clark,
  and Hajishirzi}]{khashabi-etal-2020-unifiedqa}
Daniel Khashabi, Sewon Min, Tushar Khot, Ashish Sabharwal, Oyvind Tafjord,
  Peter Clark, and Hannaneh Hajishirzi. 2020.
\newblock \href {https://doi.org/10.18653/v1/2020.findings-emnlp.171}
  {{UNIFIEDQA}: Crossing format boundaries with a single {QA} system}.
\newblock In \emph{Findings of the Association for Computational Linguistics:
  EMNLP 2020}, pages 1896--1907, Online. Association for Computational
  Linguistics.

\bibitem[{Kingma and Ba(2014)}]{kingma2014adam}
Diederik~P Kingma and Jimmy Ba. 2014.
\newblock Adam: A method for stochastic optimization.
\newblock In \emph{International Conference on Learning Representations
  (ICLR)}.

\bibitem[{Klein et~al.(2022)Klein, Hirsch, Eliav, Pyatkin, Caciularu, and
  Dagan}]{Klein2022QASemPT}
Ayal Klein, Eran Hirsch, Ron Eliav, Valentina Pyatkin, Avi Caciularu, and Ido
  Dagan. 2022.
\newblock \href {https://aclanthology.org/2022.emnlp-main.528} {{QAS}em
  parsing: Text-to-text modeling of {QA}-based semantics}.
\newblock In \emph{Proceedings of the 2022 Conference on Empirical Methods in
  Natural Language Processing}, pages 7742--7756, Abu Dhabi, United Arab
  Emirates. Association for Computational Linguistics.

\bibitem[{Klein et~al.(2020)Klein, Mamou, Pyatkin, Stepanov, He, Roth,
  Zettlemoyer, and Dagan}]{klein-etal-2020-qanom}
Ayal Klein, Jonathan Mamou, Valentina Pyatkin, Daniela Stepanov, Hangfeng He,
  Dan Roth, Luke Zettlemoyer, and Ido Dagan. 2020.
\newblock \href {https://doi.org/10.18653/v1/2020.coling-main.274} {{QAN}om:
  Question-answer driven {SRL} for nominalizations}.
\newblock In \emph{Proceedings of the 28th International Conference on
  Computational Linguistics}, pages 3069--3083, Barcelona, Spain (Online).
  International Committee on Computational Linguistics.

\bibitem[{Lewis et~al.(2020)Lewis, Liu, Goyal, Ghazvininejad, Mohamed, Levy,
  Stoyanov, and Zettlemoyer}]{lewis-etal-2020-bart}
Mike Lewis, Yinhan Liu, Naman Goyal, Marjan Ghazvininejad, Abdelrahman Mohamed,
  Omer Levy, Veselin Stoyanov, and Luke Zettlemoyer. 2020.
\newblock \href {https://doi.org/10.18653/v1/2020.acl-main.703} {{BART}:
  Denoising sequence-to-sequence pre-training for natural language generation,
  translation, and comprehension}.
\newblock In \emph{Proceedings of the 58th Annual Meeting of the Association
  for Computational Linguistics}, pages 7871--7880, Online. Association for
  Computational Linguistics.

\bibitem[{Lin and Rey(2004)}]{lin2004rouge}
C.~Lin and M.~Rey. 2004.
\newblock Looking for a few good metrics: {ROUGE} and its evaluation.
\newblock In \emph{NTCIR Workshop}.

\bibitem[{Lin(2004)}]{lin-2004-rouge}
Chin-Yew Lin. 2004.
\newblock \href {https://aclanthology.org/W04-1013} {{ROUGE}: A package for
  automatic evaluation of summaries}.
\newblock In \emph{Text Summarization Branches Out}, pages 74--81, Barcelona,
  Spain. Association for Computational Linguistics.

\bibitem[{Liu and Lapata(2019)}]{liu-lapata-2019-hierarchical}
Yang Liu and Mirella Lapata. 2019.
\newblock \href {https://doi.org/10.18653/v1/P19-1500} {Hierarchical
  transformers for multi-document summarization}.
\newblock In \emph{Proceedings of the 57th Annual Meeting of the Association
  for Computational Linguistics}, pages 5070--5081, Florence, Italy.
  Association for Computational Linguistics.

\bibitem[{Lu et~al.(2020)Lu, Dong, and Charlin}]{lu-etal-2020-multi-xscience}
Yao Lu, Yue Dong, and Laurent Charlin. 2020.
\newblock \href {https://doi.org/10.18653/v1/2020.emnlp-main.648}
  {Multi-{XS}cience: A large-scale dataset for extreme multi-document
  summarization of scientific articles}.
\newblock In \emph{Proceedings of the 2020 Conference on Empirical Methods in
  Natural Language Processing (EMNLP)}, pages 8068--8074, Online. Association
  for Computational Linguistics.

\bibitem[{Ma et~al.(2022)Ma, Zhang, Guo, Wang, and Sheng}]{10.1145/3529754}
Congbo Ma, Wei~Emma Zhang, Mingyu Guo, Hu~Wang, and Quan~Z. Sheng. 2022.
\newblock Multi-document summarization via deep learning techniques: A survey.
\newblock \emph{ACM Comput. Surv.}, 55(5).

\bibitem[{OpenAI(2023)}]{OpenAI2023GPT4TR}
OpenAI. 2023.
\newblock Gpt-4 technical report.
\newblock \emph{ArXiv}, abs/2303.08774.

\bibitem[{Ouyang et~al.(2022)Ouyang, Wu, Jiang, Almeida, Wainwright, Mishkin,
  Zhang, Agarwal, Slama, Ray, Schulman, Hilton, Kelton, Miller, Simens, Askell,
  Welinder, Christiano, Leike, and Lowe}]{NEURIPS2022_b1efde53}
Long Ouyang, Jeffrey Wu, Xu~Jiang, Diogo Almeida, Carroll Wainwright, Pamela
  Mishkin, Chong Zhang, Sandhini Agarwal, Katarina Slama, Alex Ray, John
  Schulman, Jacob Hilton, Fraser Kelton, Luke Miller, Maddie Simens, Amanda
  Askell, Peter Welinder, Paul~F Christiano, Jan Leike, and Ryan Lowe. 2022.
\newblock Training language models to follow instructions with human feedback.
\newblock In \emph{Advances in Neural Information Processing Systems
  (NeurIPS)}.

\bibitem[{Pasunuru et~al.(2021{\natexlab{a}})Pasunuru, Celikyilmaz, Galley,
  Xiong, Zhang, Bansal, and Gao}]{pasunuru2021data}
Ramakanth Pasunuru, Asli Celikyilmaz, Michel Galley, Chenyan Xiong, Yizhe
  Zhang, Mohit Bansal, and Jianfeng Gao. 2021{\natexlab{a}}.
\newblock Data augmentation for abstractive query-focused multi-document
  summarization.
\newblock In \emph{The Association for the Advancement of Artificial
  Intelligence (AAAI)}.

\bibitem[{Pasunuru et~al.(2021{\natexlab{b}})Pasunuru, Liu, Bansal, Ravi, and
  Dreyer}]{pasunuru-etal-2021-efficiently}
Ramakanth Pasunuru, Mengwen Liu, Mohit Bansal, Sujith Ravi, and Markus Dreyer.
  2021{\natexlab{b}}.
\newblock \href {https://doi.org/10.18653/v1/2021.naacl-main.380} {Efficiently
  summarizing text and graph encodings of multi-document clusters}.
\newblock In \emph{Proceedings of the 2021 Conference of the North American
  Chapter of the Association for Computational Linguistics: Human Language
  Technologies}, pages 4768--4779, Online. Association for Computational
  Linguistics.

\bibitem[{Pyatkin et~al.(2020)Pyatkin, Klein, Tsarfaty, and
  Dagan}]{pyatkin-etal-2020-qadiscourse}
Valentina Pyatkin, Ayal Klein, Reut Tsarfaty, and Ido Dagan. 2020.
\newblock \href {https://doi.org/10.18653/v1/2020.emnlp-main.224}
  {{QAD}iscourse - {D}iscourse {R}elations as {QA} {P}airs: {R}epresentation,
  {C}rowdsourcing and {B}aselines}.
\newblock In \emph{Proceedings of the 2020 Conference on Empirical Methods in
  Natural Language Processing (EMNLP)}, pages 2804--2819, Online. Association
  for Computational Linguistics.

\bibitem[{Pyatkin et~al.(2021)Pyatkin, Roit, Michael, Goldberg, Tsarfaty, and
  Dagan}]{pyatkin-etal-2021-asking}
Valentina Pyatkin, Paul Roit, Julian Michael, Yoav Goldberg, Reut Tsarfaty, and
  Ido Dagan. 2021.
\newblock \href {https://doi.org/10.18653/v1/2021.emnlp-main.108} {Asking it
  all: Generating contextualized questions for any semantic role}.
\newblock In \emph{Proceedings of the 2021 Conference on Empirical Methods in
  Natural Language Processing}, pages 1429--1441, Online and Punta Cana,
  Dominican Republic. Association for Computational Linguistics.

\bibitem[{Raffel et~al.(2020)Raffel, Shazeer, Roberts, Lee, Narang, Matena,
  Zhou, Li, and Liu}]{T5}
Colin Raffel, Noam Shazeer, Adam Roberts, Katherine Lee, Sharan Narang, Michael
  Matena, Yanqi Zhou, Wei Li, and Peter~J. Liu. 2020.
\newblock Exploring the limits of transfer learning with a unified text-to-text
  transformer.
\newblock \emph{Journal of Machine Learning Research}, 21(140):1--67.

\bibitem[{Robertson and Walker(1994)}]{robertson1994some}
Stephen~E Robertson and Steve Walker. 1994.
\newblock Some simple effective approximations to the 2-poisson model for
  probabilistic weighted retrieval.
\newblock In \emph{SIGIR’94}, pages 232--241. Springer.

\bibitem[{Tay et~al.(2021)Tay, Dehghani, Abnar, Shen, Bahri, Pham, Rao, Yang,
  Ruder, and Metzler}]{tay2021long}
Yi~Tay, Mostafa Dehghani, Samira Abnar, Yikang Shen, Dara Bahri, Philip Pham,
  Jinfeng Rao, Liu Yang, Sebastian Ruder, and Donald Metzler. 2021.
\newblock Long range arena : A benchmark for efficient transformers.
\newblock In \emph{International Conference on Learning Representations
  (ICLR)}.

\bibitem[{Tay et~al.(2022)Tay, Dehghani, Bahri, and
  Metzler}]{long-range-transformers}
Yi~Tay, Mostafa Dehghani, Dara Bahri, and Donald Metzler. 2022.
\newblock Efficient transformers: A survey.
\newblock \emph{ACM Comput. Surv.}

\bibitem[{Vaswani et~al.(2017)Vaswani, Shazeer, Parmar, Uszkoreit, Jones,
  Gomez, Kaiser, and Polosukhin}]{vaswani2017attention}
Ashish Vaswani, Noam Shazeer, Niki Parmar, Jakob Uszkoreit, Llion Jones,
  Aidan~N Gomez, {\L}ukasz Kaiser, and Illia Polosukhin. 2017.
\newblock Attention is all you need.
\newblock In \emph{Advances in Neural Information Processing Systems (NIPS)}.

\bibitem[{Wang et~al.(2020)Wang, Liu, Zheng, Qiu, and
  Huang}]{wang-etal-2020-heterogeneous}
Danqing Wang, Pengfei Liu, Yining Zheng, Xipeng Qiu, and Xuanjing Huang. 2020.
\newblock \href {https://doi.org/10.18653/v1/2020.acl-main.553} {Heterogeneous
  graph neural networks for extractive document summarization}.
\newblock In \emph{Proceedings of the 58th Annual Meeting of the Association
  for Computational Linguistics}, pages 6209--6219, Online. Association for
  Computational Linguistics.

\bibitem[{Welbl et~al.(2018)Welbl, Stenetorp, and
  Riedel}]{welbl-etal-2018-constructing}
Johannes Welbl, Pontus Stenetorp, and Sebastian Riedel. 2018.
\newblock \href {https://doi.org/10.1162/tacl_a_00021} {Constructing datasets
  for multi-hop reading comprehension across documents}.
\newblock \emph{Transactions of the Association for Computational Linguistics},
  6:287--302.

\bibitem[{Wolf et~al.(2020)Wolf, Debut, Sanh, Chaumond, Delangue, Moi, Cistac,
  Rault, Louf, Funtowicz, Davison, Shleifer, von Platen, Ma, Jernite, Plu, Xu,
  Le~Scao, Gugger, Drame, Lhoest, and Rush}]{wolf-etal-2020-transformers}
Thomas Wolf, Lysandre Debut, Victor Sanh, Julien Chaumond, Clement Delangue,
  Anthony Moi, Pierric Cistac, Tim Rault, Remi Louf, Morgan Funtowicz, Joe
  Davison, Sam Shleifer, Patrick von Platen, Clara Ma, Yacine Jernite, Julien
  Plu, Canwen Xu, Teven Le~Scao, Sylvain Gugger, Mariama Drame, Quentin Lhoest,
  and Alexander Rush. 2020.
\newblock \href {https://doi.org/10.18653/v1/2020.emnlp-demos.6} {Transformers:
  State-of-the-art natural language processing}.
\newblock In \emph{Proceedings of the 2020 Conference on Empirical Methods in
  Natural Language Processing: System Demonstrations}, pages 38--45, Online.
  Association for Computational Linguistics.

\bibitem[{Xiao et~al.(2022)Xiao, Beltagy, Carenini, and
  Cohan}]{xiao-etal-2022-primera}
Wen Xiao, Iz~Beltagy, Giuseppe Carenini, and Arman Cohan. 2022.
\newblock \href {https://doi.org/10.18653/v1/2022.acl-long.360} {{PRIMERA}:
  Pyramid-based masked sentence pre-training for multi-document summarization}.
\newblock In \emph{Proceedings of the 60th Annual Meeting of the Association
  for Computational Linguistics (Volume 1: Long Papers)}, pages 5245--5263,
  Dublin, Ireland. Association for Computational Linguistics.

\bibitem[{Xu and Lapata(2020)}]{xu-lapata-2020-coarse}
Yumo Xu and Mirella Lapata. 2020.
\newblock \href {https://doi.org/10.18653/v1/2020.emnlp-main.296}
  {Coarse-to-fine query focused multi-document summarization}.
\newblock In \emph{Proceedings of the 2020 Conference on Empirical Methods in
  Natural Language Processing (EMNLP)}, pages 3632--3645, Online. Association
  for Computational Linguistics.

\bibitem[{Yang et~al.(2018)Yang, Qi, Zhang, Bengio, Cohen, Salakhutdinov, and
  Manning}]{yang-etal-2018-hotpotqa}
Zhilin Yang, Peng Qi, Saizheng Zhang, Yoshua Bengio, William Cohen, Ruslan
  Salakhutdinov, and Christopher~D. Manning. 2018.
\newblock \href {https://doi.org/10.18653/v1/D18-1259} {{H}otpot{QA}: A dataset
  for diverse, explainable multi-hop question answering}.
\newblock In \emph{Proceedings of the 2018 Conference on Empirical Methods in
  Natural Language Processing}, pages 2369--2380, Brussels, Belgium.
  Association for Computational Linguistics.

\bibitem[{Yasunaga et~al.(2022)Yasunaga, Leskovec, and
  Liang}]{yasunaga-etal-2022-linkbert}
Michihiro Yasunaga, Jure Leskovec, and Percy Liang. 2022.
\newblock \href {https://doi.org/10.18653/v1/2022.acl-long.551} {{L}ink{BERT}:
  Pretraining language models with document links}.
\newblock In \emph{Proceedings of the 60th Annual Meeting of the Association
  for Computational Linguistics (Volume 1: Long Papers)}, pages 8003--8016,
  Dublin, Ireland. Association for Computational Linguistics.

\bibitem[{Zaheer et~al.(2020)Zaheer, Guruganesh, Dubey, Ainslie, Alberti,
  Ontanon, Pham, Ravula, Wang, Yang et~al.}]{zaheer2020bigbird}
Manzil Zaheer, Guru Guruganesh, Kumar~Avinava Dubey, Joshua Ainslie, Chris
  Alberti, Santiago Ontanon, Philip Pham, Anirudh Ravula, Qifan Wang, Li~Yang,
  et~al. 2020.
\newblock Big bird: Transformers for longer sequences.
\newblock \emph{Advances in Neural Information Processing Systems (NeurIPS)}.

\bibitem[{Zhang et~al.(2020)Zhang, Zhao, Saleh, and Liu}]{pegasus}
Jingqing Zhang, Yao Zhao, Mohammad Saleh, and Peter Liu. 2020.
\newblock {PEGASUS}: Pre-training with extracted gap-sentences for abstractive
  summarization.
\newblock In \emph{Proceedings of the International Conference on Machine
  Learning (ICML)}.

\bibitem[{Zhao et~al.(2020)Zhao, Liu, Gao, Jin, Du, Zhao, Zhang, and
  Haffari}]{10.1145/3397271.3401327}
Jinming Zhao, Ming Liu, Longxiang Gao, Yuan Jin, Lan Du, He~Zhao, He~Zhang, and
  Gholamreza Haffari. 2020.
\newblock Summpip: Unsupervised multi-document summarization with sentence
  graph compression.
\newblock In \emph{Proceedings of the International ACM SIGIR Conference on
  Research and Development in Information Retrieval (SIGIR)}.

\end{thebibliography}
\bibliographystyle{acl_natbib}

\clearpage
\appendix
\label{sec:appendix}

\section{Data Creation}
\label{app:data}
As noted in \S\ref{sec:data_creation}, we used the NewSHead corpus~\cite{newshead}. We followed the data pre-processing procedure suggested by~\citet{xiao-etal-2022-primera} which supplied each sentence in the NewSHead corpus with their \pegasus{} scores~\cite{pegasus}.\footnote{We used the publicly available data from \url{https://storage.googleapis.com/primer_summ/processed_pretraining_data.tar.gz}} 

\subsection{\qasem{} Details}
\label{subsec:qasem}
\qasem{}~\cite{Klein2022QASemPT} is a unified tool for parsing sentences into a systematic set of QAs that represent each sentence. The following three types of predication are included in this set: verbs, deverbal nominalizations, and informational discourse relations, and they represent the core units of information in a sentence. 

For producing the pre-training data for our \model{} model, we specifically targeted the verbal predicates for question-answer generation, since their corresponding training examples origin from the Question Answer driven Semantic Role Labeling (QA-SRL) dataset~\cite{he-etal-2015-question} which covers the largest part of the joint \qasem{} training data, and obtained the best empirical results during evaluation, compared to the other types (nominalizations and discourse relations). Using the QA-SRL formalism, every predicate-argument relation is labeled with a question-answer pair, and so natural language questions represent semantic roles, while answers correspond to arguments.

\qasem{} first executes sentence-level pre-processing for QA-SRL by running a part-of-speech tagger to identify verbs.\footnote{\qasem{} uses SpaCy 3.0 --- \url{https://spacy.io/}}. Then, the parser itself is based on a fine-tuned T5-small model~\cite{T5} which is given a single marked predicate in context at a time, and is trained on the task of producing the full set of question-answer pairs targeting this predicate.\footnote{The T5-based parser is trained on a multi-task text-to-text objective, using QA-SRL~\cite{he-etal-2015-question,fitzgerald-etal-2018-large}, QANOM~\cite{klein-etal-2020-qanom}, and on QADiscourse~\cite{pyatkin-etal-2020-qadiscourse}} The input sequence consists of the unique task prefix, the sentence, special markers for the target predicate, and the basic verbal-form of the predicate. The output is a set of QAs, and we select one pair according to the length of the answer (\S\ref{sec:data_creation}).  Since \qasem{} generates ``abstractive'' questions that replace arguments with placeholders, we follow \citet{pyatkin-etal-2021-asking} and use their model to convert the generated question into a more natural form, with contextualized arguments. Overall, we observed that this approach generally improves the quality of the questions, in addition to the contextualization utility. Figure~\ref{figure:qasem} shows an example from our dataset (based on a salient sentence from NewSHead~\cite{newshead}) that follows the description provided above.

\section{Pre-training Technical Details}
\label{app:pretraining}

We pretrain \model{} for a total number of 400K steps (the validation loss kept decreasing along the entire pre-training process), batch size of 16, Adam optimizer~\cite{kingma2014adam} with a learning rate of $3e-5$ and with 10k warmup steps and linear decay, all follows prior works~\cite{beltagy2020longformer,xiao-etal-2022-primera}. The pre-training process takes likely eight days on eight 48GB RTX8000 GPUs. Since the backbone of both \model{} and \primera{} is the Longformer Encoder-Decoder model (LED)~\cite{beltagy2020longformer} large version, they all have the same number of parameters (447M). LED uses a sparse local+global attention pattern in the encoder self-attention side, while using the full attention on decoder and cross-attention.

\section{Benchmarks Description}
\label{app:benchmarks}
In this section, we provide further details regarding the datasets we used for the model and baselines evaluation.

\subsection{Question Answering Benchmarks}
\label{app:qa}
We first describe in detail multi-document question answering tasks, and particularly the task of multi-hop question answering. Multi-hop question answering involves using a model to gather relevant information from multiple documents and combining it to provide the correct answer.

\paragraph{HotPotQA~\cite{yang-etal-2018-hotpotqa}.}
This question answering dataset consists of questions and 10 paragraphs from various Wikipedia documents, with two of the paragraphs containing the necessary information to correctly answer the question and eight additional paragraphs serving as distractors. The task involves identifying the correct answer span and identifying supporting evidence sentences. (For more details on the dataset, see~\citet{yang-etal-2018-hotpotqa}.)

\paragraph{WikiHop~\cite{welbl-etal-2018-constructing}.}
WikiHop is a dataset that includes a question, several potential answers (ranging from 2 to 79 options), and supporting contexts (ranging from 3 to 63 paragraphs), and the correct answer. This dataset does not provide any information about the intermediate steps required to arrive to the correct answer, so models are therefore tasked to deduce these steps based on the provided question and context.

\subsection{Multi-Document Summarization Benchmarks}
\label{app:summ}
We used \url{https://github.com/google-research/googleresearch/tree/master/rouge} for computing the \textsc{Rouge} score~\cite{lin2004rouge} with the default stemmer settings during the evaluation.

\paragraph{Multi-News~\cite{fabbri-etal-2019-multi}.}
This dataset is a collection of 56,216 pairs of news articles and professional editors-written summaries, all sourced from the web (\url{newser.com}). These pairs include trace-back links to the original documents. The authors of the dataset have also compared it to other datasets in terms of coverage, density, and compression, and found that the it is plausibly diverse compared to other similar benchmarks.

\paragraph{Multi-X-Science~\cite{lu-etal-2020-multi-xscience}.}
This dataset is sourced from Arxiv and Microsoft academic graphs, where the summaries are paragraphs of related work sections, while source documents include the abstracts of the query and referred papers. It is considered to have fewer positional and extractive biases than the Multi-News dataset, transforming it into a more challenging benchmark~\cite{10.1145/3529754} since the drawback of getting higher scores for a copied sentence at a specific position can be reduced.

\subsection{Query-Focused Multi-Document Summarization Benchmarks}
\label{app:qfsumm}
In this section, we describe the pair of datasets from \citet{pasunuru2021data} that were used in our experiments. Similarly to the multi-document summarization experiments (Appendix~\ref{app:summ}), we used \url{https://github.com/google-research/googleresearch/tree/master/rouge} for computing the \textsc{Rouge} score~\cite{lin2004rouge} with the default stemmer settings during the evaluation.

\paragraph{QmdsCnn.}
This dataset is based on the single-document CNN/Daily Mail (CNN/DM) summarizastion dataset~\cite{NIPS2015_afdec700}, where its documents are news articles available online and the summaries are their human written highlights. This dataset is transformed to multi-document one by firstly chunking the documents into small documents of paragraphs. Then, the titles of the articles serve as the queries which are fed to a BM25 search engine~\cite{robertson1994some}, that returns chunks from the entire dataset that are related to the title, and serve as the context documents.

\paragraph{QmdsIr.}
In this datasets, the authors suggested using an alternative to the queries that are based on titles of articles -- they use instead queries that are issued by actual search engine users, which is more realistic scenario for search use-cases. They collect queries and their top-10 results obtained by the Bing (\url{www.bing.com}) search engine. The target summary is derived from the answer passage, which is extracted from one of the top-ranked documents by Bing’s production QA system. Next, they omit the document that contains the answer passage from the context documents.

\section{Ablation Study Details}
\label{app:ablations}
In this section, we provide details regarding the baselines used during the input format ablation study that we conducted, and was presented in \S\ref{subsec:ablation}. 

The following list includes the detailed descriptions for all the ablations we used:

\begin{itemize}
    \item Pre-training \textit{without questions}. Following \citet{jia-etal-2022-question}, we omit the generated question, and pre-train the model to predict the answer with no visible question within the context. 
    \item Pre-training using \textit{random questions} per context documents. Given context documents, we sample a random held-out document from other clusters, and generate an unrelated question which is use for the irrelevant context. It is an alternative to using a question generated by one of the documents in the context.
    \item Pre-training using contexts \textit{with random context documents}. Following~\citet{caciularu-etal-2021-cdlm-cross}, we ablate \model{} by pre-training with random documents in the context (non-related documents), where allegedly, the model would not be capable to capture cross-document relationships properly, and under-perform on multi-document downstream tasks.
    \item Pre-training \textit{with prefixes}.  We add the \texttt{question:} and \texttt{context:} prefixes during training and inference. These should further direct the model with the locations of the question and context. While this setup slightly helps for QA, we show that for MDS, the no-prefix setup is preferable.
    \item Pre-training while \textit{placing the question before the context}. Recall that \model{} appends the question tokens to the end of the input sequence, after the context documents. Therefore, we establish a baseline for ablating this setup, and placing the question at the beginning of the input.
    \item Pre-training \textit{with question filtering}. The \qasem{} parser question generation model
    can be noisy, resulting in a question that cannot be
    answered or with an incorrect answer to a generated question. We therefore follow a recent automatic QA filtering strategy that suggests using a strong QA model to ensure that valid question-answer pairs are present in the dataset~\cite{alberti-etal-2019-synthetic,fang2020accelerating}.
    pre-training after question-answer filtering, using the strong UnifiedQA-v2 model~\cite{khashabi2022unifiedqa} that follows previous UnifiedQA~\cite{khashabi-etal-2020-unifiedqa} and trains on more supervised datasets. We took the fine-tuned BART-large~\cite{lewis-etal-2020-bart} as the question filter for a fair comparison with \qasem{}. We applied UnifiedQA-v2 over the question-context-answer triplets and took only the answerable questions according to the model, which left us with roughly 25\% of the entire pre-training data.
    \item Pre-training \textit{without generating the salient sentence}. Recall that we task \model{} to generate the salient sentence which was used to produce the question and answer. This should enable the model to generate longer sequences and improve the coping mechanism, which is useful for tasks such as summarization. This hypothesis is assessed by executing the same pre-training procedure but without generating the salient sentence -- only the answer of the generated question.
    \item Using alternative QA generators from recent related works. We pre-train a model based on the QAs generated by two QA generators, based on the BART-large model~\cite{lewis-etal-2020-bart}: The first is taken from~\citet{jia-etal-2022-question}\footnote{Since this model was not publicly available, we reproduced this model and fine-tuned using the HuggingFace package~\cite{wolf-etal-2020-transformers}.}, which trained a model over the data from the MRQA 2019 Shared Task~\cite{fisch-etal-2019-mrqa} and the second is the QA generator from~\cite{khashabi2022unifiedqa} which was trained on eight different QA benchmarks (see full list and references in~\citet[Appendix A]{khashabi2022unifiedqa}).
    \item Additional pre-training for \primera{}~\cite{xiao-etal-2022-primera} -- We resume the pre-training of the 100k publicly released checkpoint of \primera{}, and pre-train for an additional number of 300k steps (using the same pre-training format and procedure described in~\citet{xiao-etal-2022-primera}), to reach the number of steps used for pre-training \model{} and its ablations described above.
\end{itemize}

\section{API-Based Models Prompting Details}
\label{sec:prompt}

We manually explored several prompts for the GPT-3.5 and GPT-4 chat API-based models, and proceeded with the one that appeared to be the most effective for zero-shot multi-document summarization, as follows.

Per a Multi-News example where we are given $k$ context documents $D_1,D_2,\ldots,D_k$, we prompt each model to provide an summary using the system format: \\
\texttt{``You are a helpful assistant that summarizes important information from multiple documents.''},\\

and the user format: \\
\texttt{``Summarize the following documents into a single summary:} \\
\texttt{Document 1: $D_1$}\\
\texttt{Document 2: $D_2$}\\
$\vdots$\\
\texttt{Document k: $D_k$''}\\

\section{System Summary Examples of GPT-3 and \model{}}
\label{sec:summ_examples}

In Table \ref{table:examples}, we include three examples of system summaries produced by GPT-3.5 and \model{}, as well as the corresponding reference (ground-truth) summary. In general, \model{}'s summaries are more concise, include less redundant information, do not include anecdotal information, and overall were preferred by the human evaluators.

\begin{table*}[ht]
\setlength{\belowcaptionskip}{-10pt}
\setlength\tabcolsep{0.6pt}
\tiny
\begin{center}
\begin{tabular}{ p{0.28\linewidth} p{0.28\linewidth} p{0.28\linewidth}}
\textbf{Reference Ground-Truth Summary} & \textbf{GPT-3.5} & \textbf{\model{}} \\ 
\toprule
\texttt{You may have heard the happy news: Prince William and Kate Middleton are the proud parents of a 3-month-old cocker spaniel. However, if you were hoping to find out what they're calling the puppy, prepare to be disappointed. The Duke and Duchess of Cambridge have strictly instructed aides not to reveal his name, the Daily Mail reports. Says a spokesperson, ``He is a private pet and they do not want his name to be made public although the couple are happy to confirm that they do, indeed, have a new dog.'' Click for pictures of the ``private pet.''}
&
\texttt{Prince William and Duchess Kate Middleton have adopted a cocker spaniel puppy, which is a few months old and the son of Kate's mother's dog, Ella. The couple initially had concerns about taking care of a pet, but quickly fell in love with the puppy and decided to keep him. The Duke and Duchess of Cambridge got the male cocker spaniel in early December and have been settling him in at their rented farmhouse in North Wales. They have chosen to keep the name of their pet private. The new dog is the couple's first together and replaces Prince William's black Labrador, Widgeon, who died about two years ago.
}
& \texttt{Prince William and wife Kate Middleton have adopted a new addition to the family: a cocker spaniel puppy. The Telegraph reports that the couple has adopted a baby boy, but it's not a baby. The puppy is just a few months old and is the son of Kate's mother's dog, Ella. ``William and Catherine fell in love with the pup instantly and it wasn't long before they decided to keep him,'' a palace aide tells US Weekly. ``He's now part of the royal fold.''}
 \\
\midrule
\texttt{A rush-hour collision between a Chicago Transit Authority bus and several other vehicles yesterday left one person dead and at least eight others injured, one of them critically, authorities say. The accident occurred around 6pm in the north Loop. Authorities say the articulated Route 148 Clarendon/Michigan Express bus collided with at least three other vehicles at Michigan Avenue and Lake Street. The bus went onto the sidewalk, and at one point a pedestrian was pinned underneath. She was taken away covered in a sheet, a witness who ran to help tells the Chicago Tribune. NBC Chicago describes the fatality as a 51-year-old woman. The driver, who was treated for non-life-threatening injuries, was the only person on the bus, and investigators are looking at video from a camera that records the interior of the bus.}
&
\texttt{A Chicago Transit Authority bus was involved in a serious crash during rush hour, resulting in one fatality and eight injuries. The bus collided with several other vehicles at North Michigan Avenue and East Lake Street. The bus driver has been cited for failing to stop at a red light and for ``failure to exercise due care.'' The accident is still under investigation. The deceased has been identified as 51-year-old Aimee Coath of Flossmoor. The eight other individuals, including the bus driver, were hospitalized with non-life-threatening injuries.}
&
\texttt{A bus crash in downtown Chicago last night left one person dead and eight others injured, including the bus driver, at least 10 ambulances were called to the scene, reports NBC Chicago. The fatality has been identified as 51-year-old Aimee Coath of Flossmoor, reports the Chicago Tribune. Coath was the only person on the Chicago Transit Authority bus at the time of the crash.
}\\
 \\
\midrule
\texttt{Geez, the French are even sophisticated while performing wanton acts of destruction. The Verge reports a young man was caught on video calmly and methodically wrecking up an Apple Store in France over a refund disagreement. The man used a steel ball—apparently the kind used in a French lawn game—to break at least 10 iPhones and a MacBook Air, one at a time, before being arrested outside the store. ``Apple is a company that violated European consumers' rights,'' the Daily Dot quotes the man as saying in French during his iPhone smashing. ``They refused to reimburse me. I told them: 'Give me my money back.' They said no. So you know what's happening? This is happening!''}
&
\texttt{An Apple Store in Dijon, France was vandalized by an irate customer who used a steel ball to smash iPhones, MacBooks, and iPads. According to reports, the customer was in a dispute with Apple over a refund and claimed that the company violated European consumers' rights. He was eventually apprehended by security and arrested after causing significant damage to the store.}
&
\texttt{A video of an angry man destroying everything in a French Apple Store is making the rounds on the Internet is making headlines, and it's not for the first time. The video shows a man hurling a steel ball through a store's windows, smashing everything in sight, and then calmly waiting for security to come and stop him, reports the BBC. The man, who is in his 20s, is identified as a French citizen who lives in the Paris suburb of Montpellier. He was caught on surveillance video at the store on Wednesday.
}\\
\bottomrule
\end{tabular}
\end{center}
\caption{The system summaries and reference summary of three document clusters in Multi-News.
}
\label{table:examples}
\end{table*}

\section{List of Software and Data Licences Used  in this Work}

Our code will be released and licensed under the Apache License 2.0 license. Our framework dependencies are: 
\begin{itemize}
    \item \primera{}: \url{https://github.com/allenai/PRIMER/blob/main/LICENSE}, under an Apache License 2.0.
    \item LongT5: \url{https://github.com/google-research/longt5/blob/master/LICENSE}, under an Apache License 2.0.
    \item NewSHead: \url{https://github.com/google-research-datasets/NewSHead}, Misc.
    \item QmdsCnnIr: \url{https://github.com/ramakanth-pasunuru/QmdsCnnIr}, Misc.
    \item Multi-XScience: \url{https://github.com/yaolu/Multi-XScience/blob/master/LICENSE}, under a MIT License.
    \item Multi-News: \url{https://github.com/Alex-Fabbri/Multi-News/blob/master/LICENSE.txt}, Misc.
    \item HotpotQA: \url{https://hotpotqa.github.io}, under a CC BY-SA License 4.0.
    \item WikiHop: \url{https://qangaroo.cs.ucl.ac.uk/}, under a CC BY-SA License 3.0.
    \item Huggingface Transformers: \url{https://github.com/huggingface/transformers/blob/master/LICENSE}, under an Apache License 2.0.
    \item HuggingFace Datasets: \url{https://github.com/huggingface/datasets/blob/master/LICENSE}, under an Apache License 2.0.
    \item Huggingface Evaluate: \url{https://github.com/huggingface/evaluate/blob/main/LICENSE}, under an Apache License 2.0.
    \item Pytorch: \url{https://github.com/pytorch/pytorch/blob/master/LICENSE}, Misc.
    \item Pytorch Lightning: \url{https://github.com/PyTorchLightning/pytorch-lightning/blob/master/LICENSE}, under an Apache License 2.0.
    \item Longformer: \url{https://github.com/allenai/longformer/blob/master/LICENSE}, under an Apache License 2.0.
    \item UnifiedQA: \url{https://github.com/allenai/unifiedqa/blob/master/LICENSE}, under an Apache License 2.0.
    \item \textsc{Rouge}: \url{https://github.com/google-research/google-research/tree/master/rouge}, under an Apache License 2.0.
    \item spaCy: \url{https://github.com/explosion/spaCy/blob/master/LICENSE}, under a MIT License.
    \item NLTK: \url{https://github.com/nltk/nltk}, under an Apache License 2.0.
    \item NumPy: \url{https://github.com/numpy/numpy/blob/main/LICENSE.txt}, under a BSD 3-Clause ``New'' or ``Revised'' License.
    \item seaborn: \url{https://github.com/mwaskom/seaborn/blob/master/LICENSE.md}, under a BSD 3-Clause ``New'' or ``Revised'' License.
    \item openai: \url{https://github.com/openai/openai-python/blob/main/LICENSE}, under a MIT License.

\end{itemize}

\end{document}